\documentclass[journal]{IEEEtran}
\usepackage{amsmath,amsfonts}
\usepackage{algorithmic}
\usepackage{algorithm}
\usepackage{array}
\usepackage{amssymb}
\usepackage{subcaption} 
\usepackage{booktabs}
\usepackage{multirow}
\usepackage{makecell}
\usepackage{stfloats}
\usepackage{url}
\usepackage{verbatim}
\usepackage{tabularx}
\usepackage{cite}
\usepackage{float}
\usepackage{threeparttable}
\usepackage{pifont}
\usepackage{xcolor}
\usepackage[colorlinks=true,linkcolor=blue,citecolor=blue,urlcolor=blue]{hyperref}
\usepackage{perpage} 
\MakePerPage{footnote} 
\usepackage{graphicx}
\usepackage{caption}
\begin{document}
\title{UCS: Universal Model for Curvilinear Structure Segmentation}
\author{Kai Zhu, Li Chen, Dianshuo Li, Yunxiang Cao, Jun Cheng, \IEEEmembership{Senior Member, IEEE}
\thanks{This paper was supported by the National Natural Science Foundation of China (62271359) and the Agency for Science, Technology and Research (A*STAR) under its MTC Programmatic Funds (Grant No. M23L7b0021).}
\thanks{Kai Zhu, Li Chen, Dianshuo Li, Yunxiang Cao, are with School of Computer Science and Technology, Wuhan University of Science and Technology, Wuhan, China, and Hubei Province Key Laboratory of Intelligent Information Processing and Real-time Industrial System, Wuhan University of Science and Technology, Wuhan, 430065, China (e-mail: lidianshuo@wust.edu.cn; chenli@wust.edu.cn; cyx529630@gmail.com; zhukai@wust.edu.cn).}
\thanks{Jun Cheng is with the Institute for Infocomm Research (I$^2$R), Agency for Science, Technology and Research (A*STAR), 1 Fusionopolis Way, \#21-01, Connexis South Tower, Singapore 138632, Republic of Singapore (e-mail: cheng\_jun@a-star.edu.sg).}}

\markboth{Journal of \LaTeX\ Class Files,~Vol.~14, No.~8, August~2021}%
{Shell \MakeLowercase{\textit{et al.}}: A Sample Article Using IEEEtran.cls for IEEE Journals}

\IEEEpubid{0000--0000/00\$00.00~\copyright~2021 IEEE}

\maketitle
\begin{abstract}
Curvilinear structure segmentation (CSS) is essential in various domains, including medical imaging, landscape analysis, industrial surface inspection, and plant analysis. While existing methods achieve high performance within specific domains, their generalizability is limited. On the other hand, large-scale models such as Segment Anything Model (SAM) exhibit strong generalization but are not optimized for curvilinear structures. Existing adaptations of SAM primarily focus on general object segmentation and lack specialized design for CSS tasks. To bridge this gap, we propose the Universal Curvilinear structure Segmentation (UCS) model, which adapts SAM to CSS tasks while further enhancing its cross-domain generalization. UCS features a novel encoder architecture integrating a pretrained SAM encoder with two innovations: a Sparse Adapter, strategically inserted to inherit the pre-trained SAM encoder's generalization capability while minimizing the number of fine-tuning parameters, and a Prompt Generation module, which leverages Fast Fourier Transform with a high-pass filter to generate curve-specific prompts. Furthermore, the UCS incorporates a mask decoder that eliminates reliance on manual interaction through a dual-compression module: a Hierarchical Feature Compression module, which aggregates the outputs of the sampled encoder to enhance detail preservation, and a Guidance Feature Compression module, which extracts and compresses image-driven guidance features. Evaluated on a comprehensive multi-domain dataset, including an in-house dataset covering eight natural curvilinear structures, UCS demonstrates state-of-the-art generalization and open-set segmentation performance across medical, engineering, natural, and plant imagery, establishing a new benchmark for universal CSS. The source code is available at \url{https://github.com/kylechuuuuu/UCS}.
\end{abstract}


\section{Introduction}
\IEEEPARstart{C}URVILINEAR structures~\cite{1995curve,2001curves}, characterized by intertwined slender structures with varying lengths and thicknesses, pose challenges in segmentation tasks, requiring precise extraction of their locations and distributions for further analysis and processing. 

\begin{figure}[ht]
    \centering
    \includegraphics[width=\linewidth]{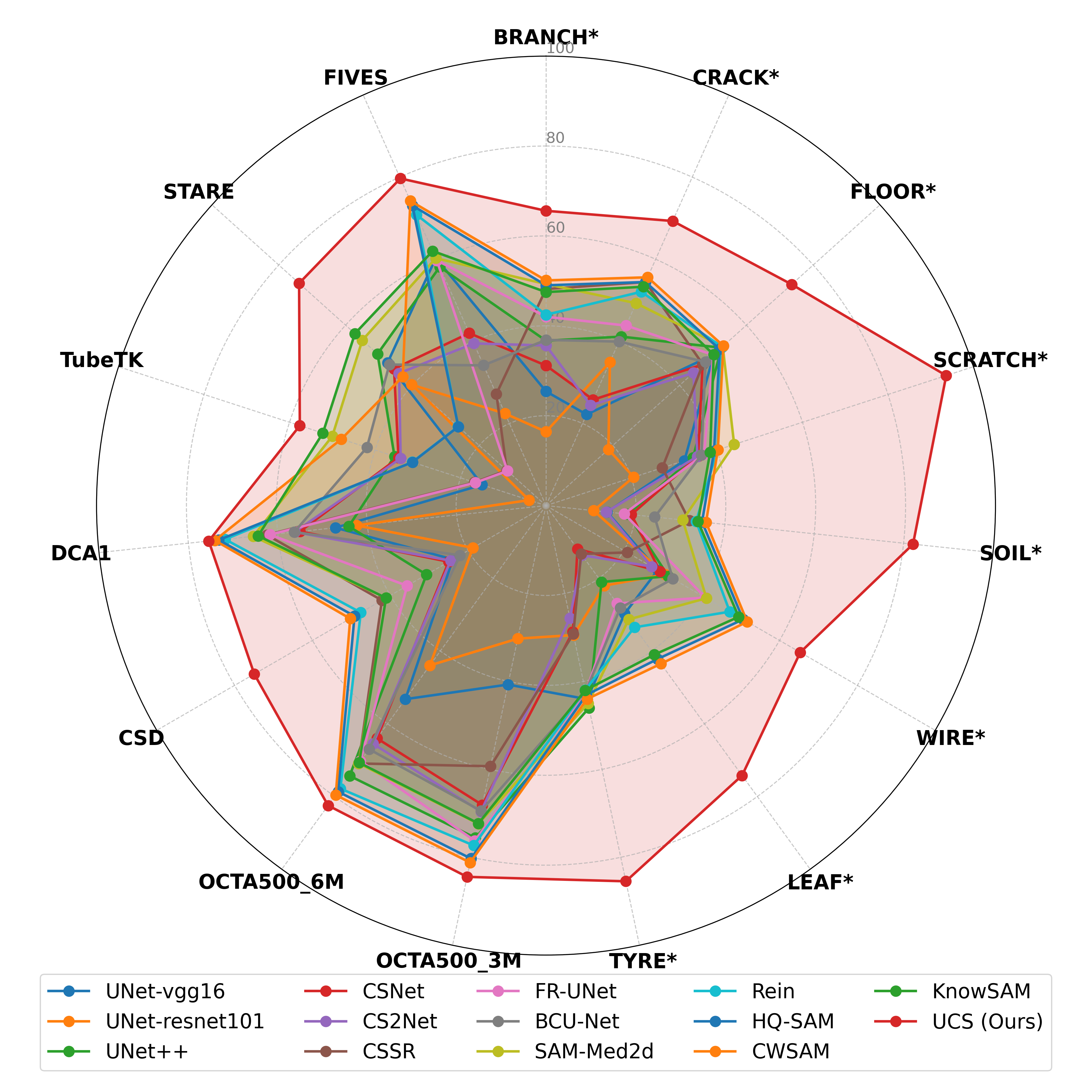}
    \caption{Comparison of F1-score across diverse segmentation models and datasets, where $^{\ast}$ denotes in-house dataset.}
    \label{fig: radar}
\end{figure}

Widely present across diverse domains, their accurate segmentation is essential for many applications. In medical images~\cite{3dmedical, Retinal}, the accurate extraction of blood vessels facilitates the early diagnosis and monitoring of vascular diseases, directly impacting patient care and treatment planning. In pavement engineering~\cite{shi2016automatic, yang2019feature, DeeperCrack, CSSR}, detecting cracks ensures structural integrity and operational safety, minimizing the risk and reducing maintenance costs. Similarly, in plant biology~\cite{leaf3}, the accurate segmentation of leaf venation enables an understanding of the adaptive and ecological significance of leaf vein network structure, representing an essential component of plant phenotyping in ecological and genetic studies~\cite{leaf2}.
\IEEEpubidadjcol

\begin{figure*}[ht]
\setlength{\tabcolsep}{0.1pt}
  \subfloat[{\normalfont Inputs}]{
     \centering
     \includegraphics[width=0.50\textwidth]{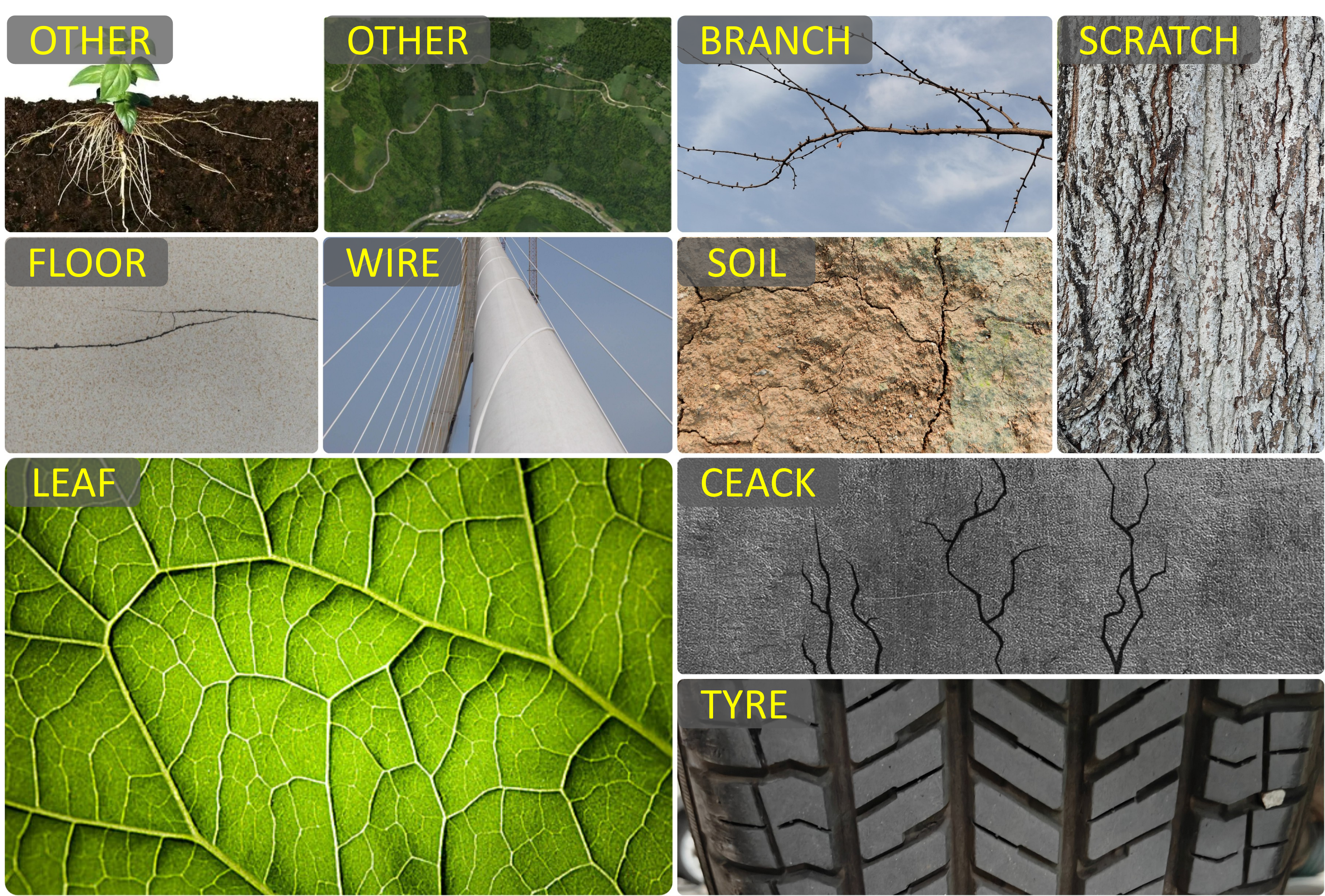}
  }
  \subfloat[{\normalfont Results}]{
     \centering
     \includegraphics[width=0.50\textwidth]{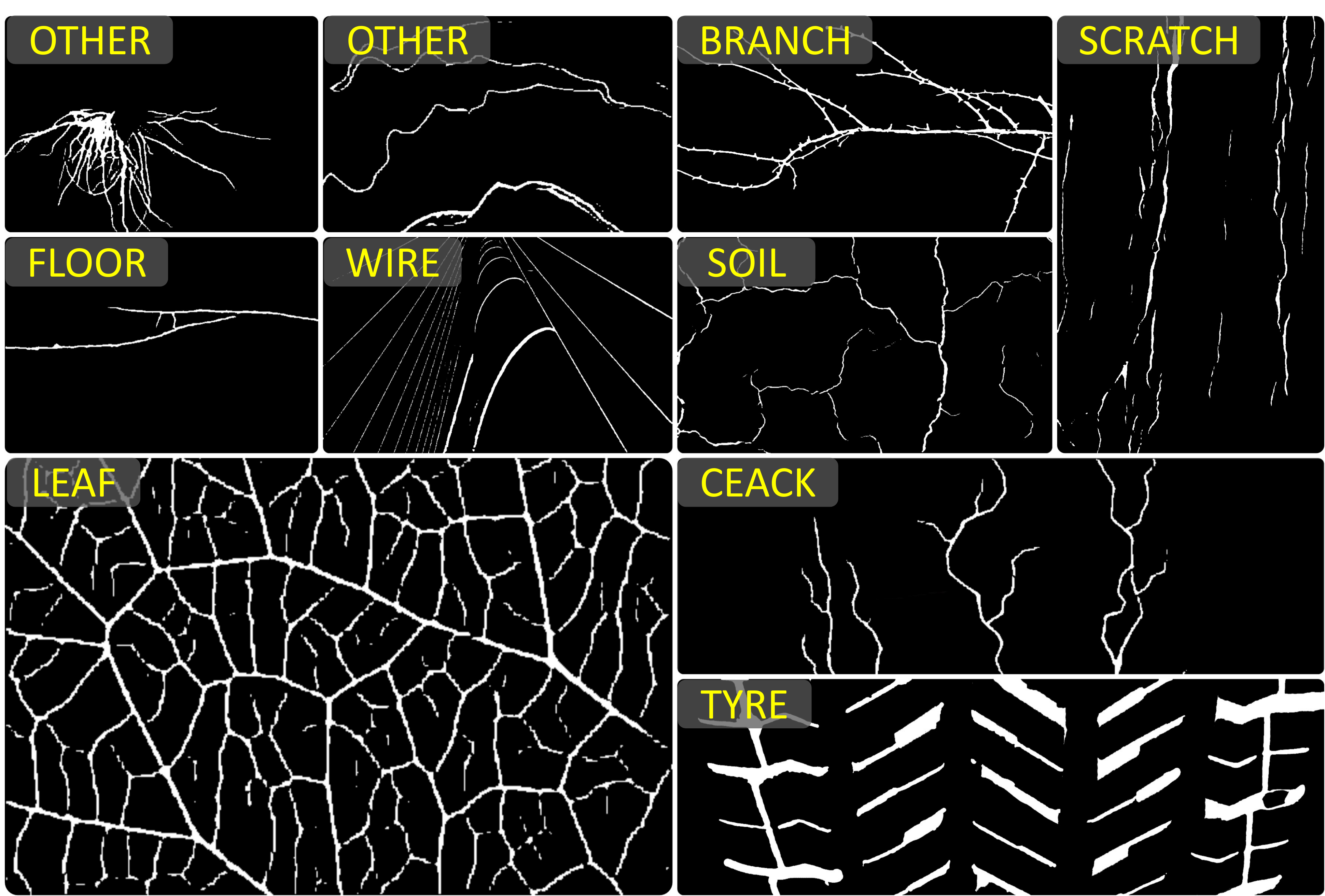}
  }
\caption{
Visualization results of curvilinear segmentation by UCS across various scenarios. These images are selected from our in-house dataset and represent key examples from eight scene categories. The (a) shows the input images, while the (b) presents the corresponding segmentation results.
}
\label{demoresult}
\end{figure*}

Although curvilinear structures across various domains share similar geometric characteristics, their representations exhibit substantial variations depending on the specific domain. This domain gap poses a considerable challenge for curvilinear structure segmentation. In this work, we use the term universal in a task-specific sense: a single model that can be trained once and then applied, without architecture changes, to segment curvilinear structures across heterogeneous domains (e.g., medical, engineering, natural, and plant imagery) and imaging modalities (e.g., fundus, OCTA, X-ray, RGB) under both seen and unseen scenarios. Developing such a universal model that can overcome cross-domain limitations is essential. It would enable cross-domain applications and drive innovation in transfer learning, open-set segmentation, and resource-efficient model deployment, while also highlighting the remaining challenges in extending universality to even broader modalities such as MRI or CT.

Given the need for such a universal model, we first examined the characteristics and limitations of traditional curvilinear structure segmentation approaches. These methods~\cite{bloodsegment, 2015unet, 9733198, CSSR, unet++, FR-UNet, BCU} typically require domain-specific training, and the domain gap between datasets limits their direct application to novel domains, impeding their wider adoption for curvilinear structure segmentation across diverse fields. Additionally, the trend of large-scale models has had a profound impact across the field of AI. The emergence of general-purpose AI models, such as GPT-4V~\cite{gpt4-v}, DINO~\cite{maskdino}, SegGPT~\cite{seggpt}, APE~\cite{APE}, SAM~\cite{SAM}, facilitates the use of foundation model to solve multiple different downstream tasks. 

The SAM\cite{SAM} marks a significant advancement in extending the capabilities of segmentation models, offering powerful zero-shot capabilities and flexible prompting mechanisms. Thus, many works~\cite{eviprompt, medsam, sam-med2d, SAMinHQ, SAMforremote, samadapter} have explored the application of SAM in several research fields. MedSAM~\cite{medsam} and SAM-Med2d~\cite{sam-med2d} have explored applying SAM to medical image segmentation. SAM for Remote Sensing~\cite{SAMforremote} has adapted SAM's prior knowledge to the remote sensing domain. Although these methods have achieved promising results in their respective domains, these methods are domain-specific and lack multi-domain applicability. 

In contrast, we propose UCS, which leverages the pretrained knowledge of SAM to enable cross-domain curvilinear structure segmentation. Extensive experiments demonstrate that UCS achieves remarkable domain adaptation capability through fine-tuning on limited data (3,244 images), successfully segmenting curvilinear structures in both seen and unseen domains. As illustrated in Fig.~\ref{demoresult}, UCS exhibits exceptional zero-shot generalization across images from diverse scenes. In summary, our main contributions are summarized as follows:
\begin{enumerate}
\item
We introduce UCS, the first universal framework for cross-domain curvilinear structure segmentation. It achieves state-of-the-art zero-shot performance and robust generalization to unseen domains with fine-tuning on limited data.
\item
We introduce an encoder built upon a pretrained image encoder, featuring two key innovations: (1) Different from one-block, one-adapter idea, the adapters are sparsely inserted at the encoder blocks. This Sparse Adapter inherits the SAM encoder's generalization capabilities and minimizes the number of fine-tuning parameters. (2) The Prompt Generation (PG) module integrated a Fast Fourier Transform with a high-pass filter to extract high-frequency responses via a residual connection. It's designed to generate domain-adaptive initial prompts by exploiting frequency-domain representations of the input data.
\item
We propose a mask decoder that enhances feature representation and eliminates reliance on manual interaction through a dual-compression module: (1) Hierarchical Feature Compression (HFC) module compresses the sampled encoder outputs with an interval of 6 and adds them to the decoder layer outputs. It provides multi-layered information, enriching the feature space available to the decoder. (2) The Guidance Feature Compression (GFC) module employed Principal Component Analysis (PCA) to extract the guidance features of the decoder layer. It enables image-based automatic segmentation and eliminates the need for manual prompts.
\item
Extensive experiments demonstrate that UCS achieves excellent performance across multiple in-house and public benchmark datasets, confirming its robustness and cross-domain generalization.
\end{enumerate}

\section{Related Work}
\subsection{Curvilinear Structure Segmentation with CNN}
Since the invention of U-Net~\cite{2015unet}, designed to integrate low-level and high-level information for image segmentation, numerous U-Net variants have been proposed. Among these, the FR-UNet~\cite{FR-UNet} enhances the connectivity of U-Net segmentation through a multi-resolution convolution interactive mechanism and a dual-threshold iterative algorithm. 

Building upon U-Net, several CNN-based models have been optimized to achieve more accurate segmentation. Wang et al.~\cite{CAS} integrated VGG into U-Net and enhanced the performance of curve segmentation by cyclically sampling patches of varying sizes from the images. Xu et al.~\cite{Semicurv} proposed an N-pair consistency loss function and introduced 
geometric transformation data augmentation to facilitate semi-supervised learning for curvilinear segmentation tasks.

In the medical field, extensive research has focused on developing more accurate vessel segmentation methods. Zhang et al.~\cite{10550146} introduces a unified multi-task learning framework for joint segmentation and registration, effectively enhancing the accuracy and robustness of vessel segmentation and registration through a bi-fusion of structure and deformation at multiple scales. VSR-Net~\cite{VSR-Net} leverages CNN for initial coarse segmentation of vessel-like structures, followed by a graph clustering approach to dynamically rehabilitate subsection ruptures, thereby enhancing topological integrity and segmentation quality. Zhou et al.~\cite{MaskVSC} proposed a backbone-agnostic MaskVSC method to reconstruct the retinal vascular network by simulating missing sections of blood vessels and using this simulation to train the model to predict the missing parts and their connections.

In the crack segmentation field, DeepCrack~\cite{DeepCD} introduced a deep convolutional neural network that optimizes crack segmentation through hierarchical convolutional fusion of multi-scale deep features. Crack Segmentation with Super-Resolution~\cite{CSSR} jointly learns semantic segmentation and super-resolution tasks to segment cracks from low-resolution images. By incorporating boundary combo loss to address the class imbalance, CSSR achieves crack segmentation while enhancing image resolution. SFIAN~\cite{SFIAN} effectively models irregular crack objects through selective feature fusion and irregularity-aware mechanisms, achieving pavement crack detection.

\subsection{Curvilinear Structure Segmentation with Transformer}
The emergence of Transformer architectures has demonstrated capabilities in modeling long-range dependencies, offering promising potential to overcome the receptive field limitations of CNN~\cite{attentiong}. This characteristic has led researchers to explore Transformer architectures for curvilinear segmentation tasks.

In retinal vascular segmentation, the Group Transformer Network~\cite{Gtunet} includes a set of bottleneck structures that reduce the computational complexity of the Transformer, facilitating the combination of CNN and Transformer. Yu et al.~\cite{yu} first predicted the gamma values through CNN to adjust the intensity distribution of retinal images and the channel-attention Vision Transformer (ViT) enhanced edge feature maps both spatially and channel-wise. TUnet-LBF~\cite{TUnet-LBF} integrates Transformer Unet and local binary energy function model for coarse to fine segmentation of retinal vessels. The Stimulus-Guided Adaptive Transformer Network~\cite{bloodsegment} introduces a hybrid CNN-Transformer architecture with the Stimulus-Guided Adaptive Pooling Transformer to enhance global context capture and suppress redundant information. This approach effectively distinguishes blood vessels from lesions and handles blurred regions in fundus images. In crack segmentation, Qi et al.~\cite{qi} introduced an end-to-end model based on ViT and level set theory for segmenting bridge pavement defects, achieving precise segmentation by merging the outputs of two parallel decoders. 

\begin{figure*}[t] 
\centering
\includegraphics[width=0.9\textwidth]{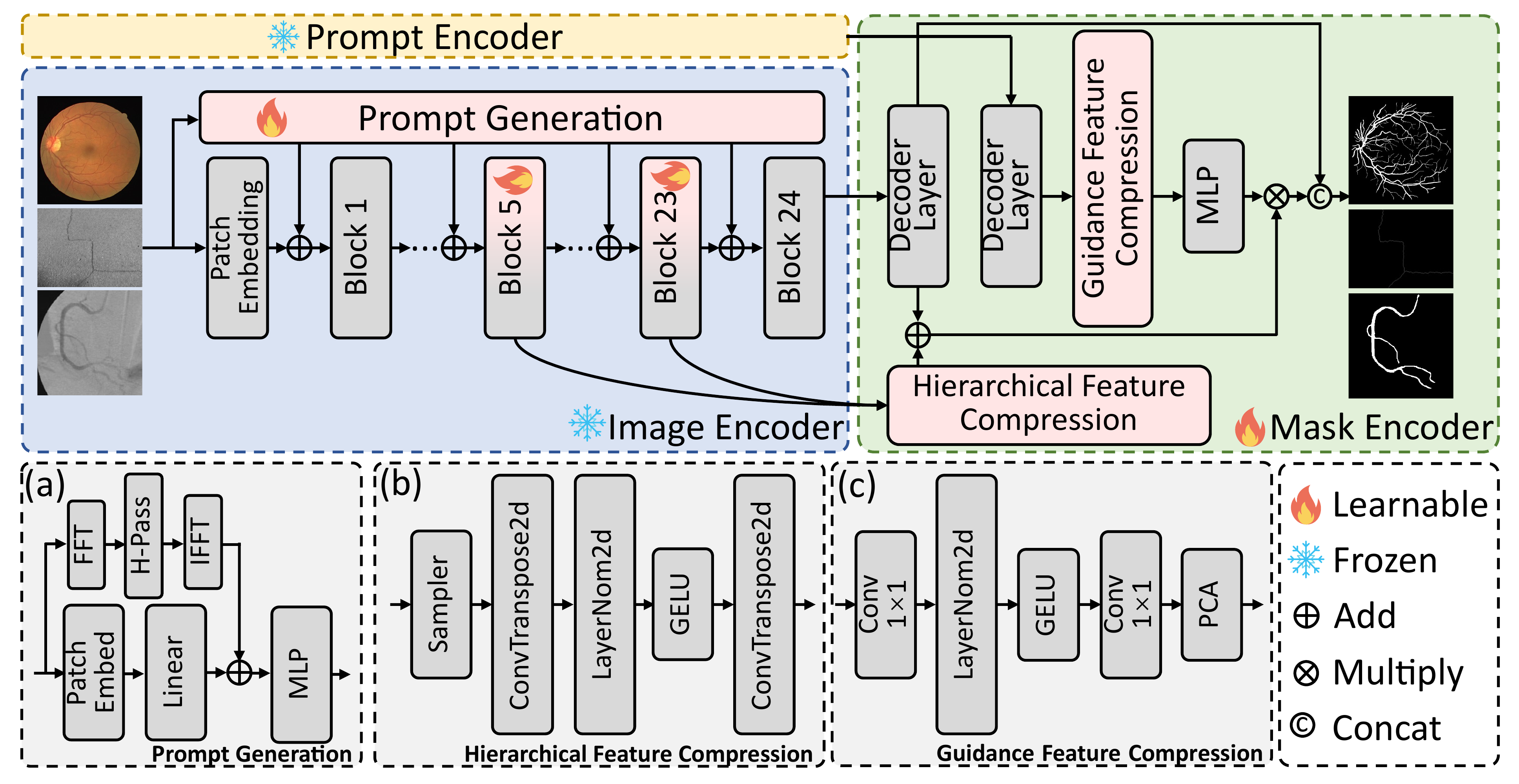} 
\caption{The proposed UCS model architecture. We freeze the image encoder and insert a learnable Sparse Adapter (SA) in specific encoder blocks, namely the 5th, 11th, 17th, and 23rd blocks, as shown in (a). The Prompt Generation (PG) module, illustrated in (b), integrates a Fast Fourier Transform with a high-pass filter to transform high frequency into dynamic prompts using a residual connection. The Hierarchical Feature Compression (HFC) module, shown in (c), compresses the outputs of the sampled encoders and adds them to the decoder layer. The Guidance Feature Compression (GFC) extracts image-driven guidance features. During training, the pretrained weights of the SAM encoder are frozen, and the rest of the model is trained.}
\label{UCS} 
\end{figure*}

\subsection{Segmentation with Large-Scale Model}
While CNN-based methods and Transformer hybrid models excel in their trained domains, the recent push towards universal models signals a shift towards large-scale model architectures. 
Thus, developing vision systems that can transfer to new concepts or domains has become a vital research topic~\cite{APE}. Many open-vocabulary models~\cite{open2022, openvocab2023, Regin} leverage large pretrained multimodal models (e.g., CLIP~\cite{clip}) to extract or transfer visual semantic knowledge. Beyond utilizing these foundational models, approaches like DenseCLIP~\cite{Denseclip} and GroupViT~\cite{Groupvit} demonstrate that fine-tuning these models or training them from scratch can also yield superior zero-shot performance. 

The SAM marks a significant advancement in extending the capabilities of segmentation models. As a result, SAM has become a popular baseline in the segmentation domain. Several works have investigated the application and adaptation of SAM to different domains. SAM-Med2d~\cite{sam-med2d} bridges the domain gap between natural and medical images by fine-tuning SAM on a large-scale medical dataset with 4.6 million images and 19.7 million masks. EviPrompt~\cite{eviprompt} takes a training-free approach, using a single reference image-annotation pair to minimize labeling and computational costs. Both explore distinct strategies for adapting SAM to medical image segmentation. HQ-SAM~\cite{SAMinHQ} focuses on improving SAM's capability for fine-grained segmentation by introducing a learnable, high-quality output token. This token is injected into SAM's mask decoder and designed to predict high-quality masks, enhancing its ability to handle detailed segmentation tasks. Building on SAM's strengths, we aim to address these gaps and explore its application for curvilinear structure segmentation across diverse domains.

\section{Method}
This section presents the proposed universal curvilinear structure segmentation model, UCS. As illustrated in Fig.~\ref{UCS}, UCS comprises two primary components: a SAM-based image encoder and a mask decoder. To achieve this, we first introduce the Sparse Adapter (SA) module, which is sparsely inserted at the 5\(th\), 11\(th\), 17\(th\), 23\(rd\) encoder blocks, inheriting the SAM encoder's generalization capabilities and minimizing the number of fine-tuning parameters. Furthermore, the PG module implements an FFT with high-pass filter to supply curve-specific prompts to each encoder block (Sec.~\ref{encoder}). Secondly, we propose the HFC and GFC module, using sample strategy and PCA to supply multi-layered and extract the guidance information (Sec.~\ref{decoder}). Finally, we design the loss functions to supervise the training of UCS (Sec.~\ref{loss}).

\begin{figure}[t]
 \centering
\subfloat[]{
\includegraphics[width=0.235\columnwidth]{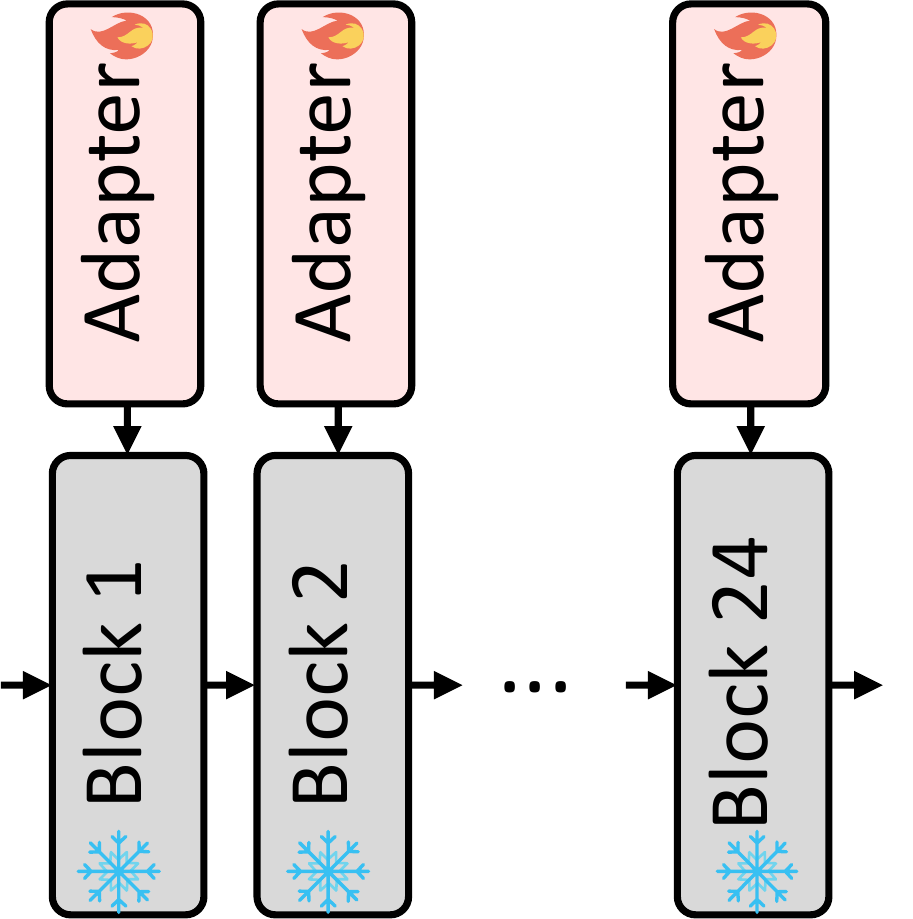}
}
\hspace{0.5em}
{\vrule width 0.5pt height 0.27\columnwidth}%
\subfloat[]{
\includegraphics[width=0.47\columnwidth]{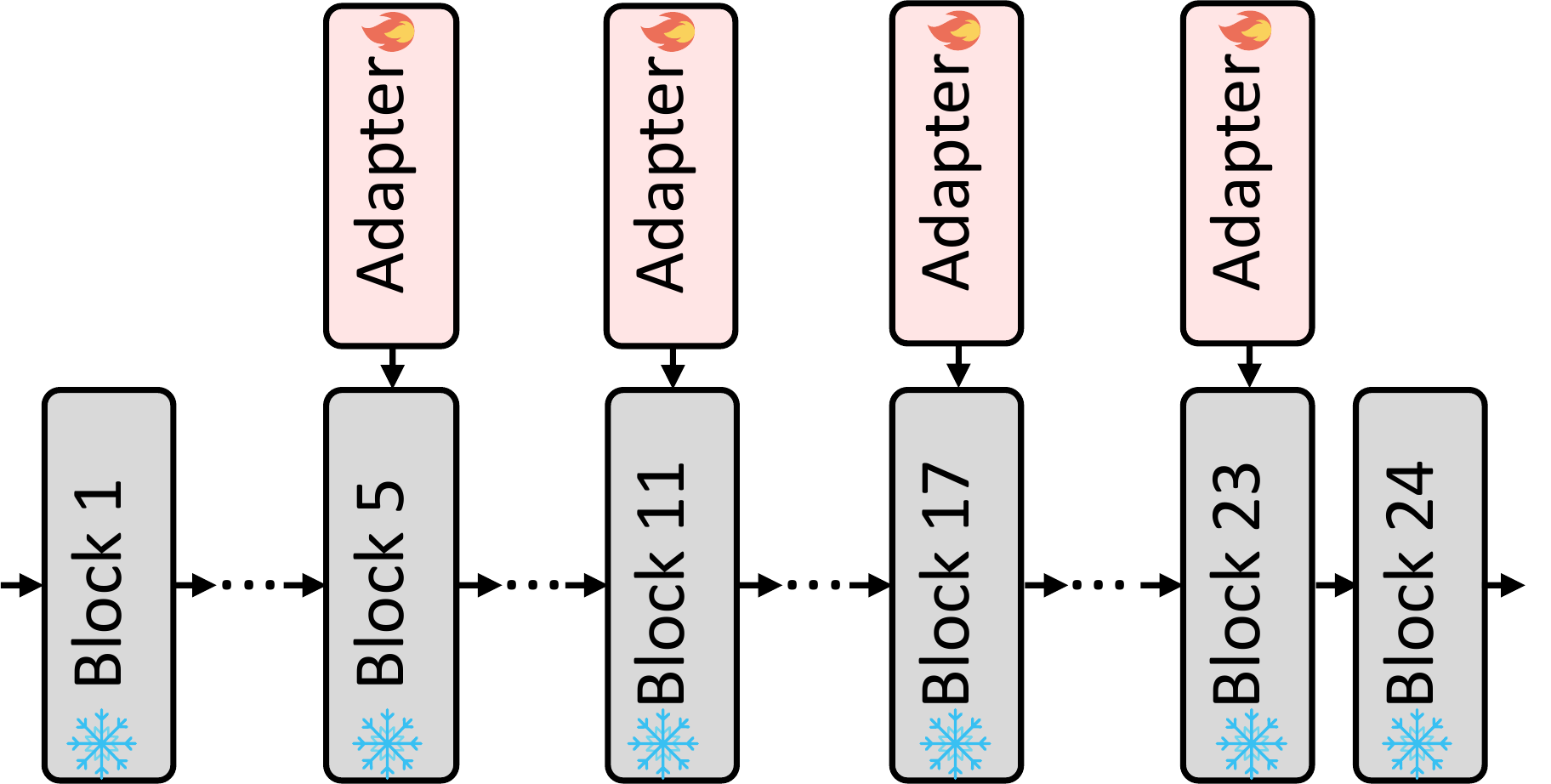}
}
\caption{(a) Previous adapter approach (b) Our Sparse Adapter.}
\label{fig: sparse}
\end{figure}

\subsection{Image Encoder}\label{encoder}
The refinement of the image encoder incorporates two key components: the Sparse Adapter module and the Prompt Generation module.

\subsubsection{\textbf{Sparse Adapter}}
Adapter has proven to be an effective strategy for fine-tuning large models for downstream tasks~\cite{adapter, adapter1}. By eliminating the need for full-parameter training, they preserve the model's original knowledge and prevent catastrophic forgetting. Existing adapters are typically designed for single-domain adaptation, limiting their ability to generalize across diverse domains. For example, SAM-Med2D~\cite{sam-med2d} improves segmentation performance but is limited to the medical domain. SAM-Adapter~\cite{samadapter} shows performance gains in shadow and camouflaged segmentation based on datasets from the same domain as the fine-tuning data.

\begin{figure}[t]
    \centering
    \includegraphics[width=0.55\columnwidth]{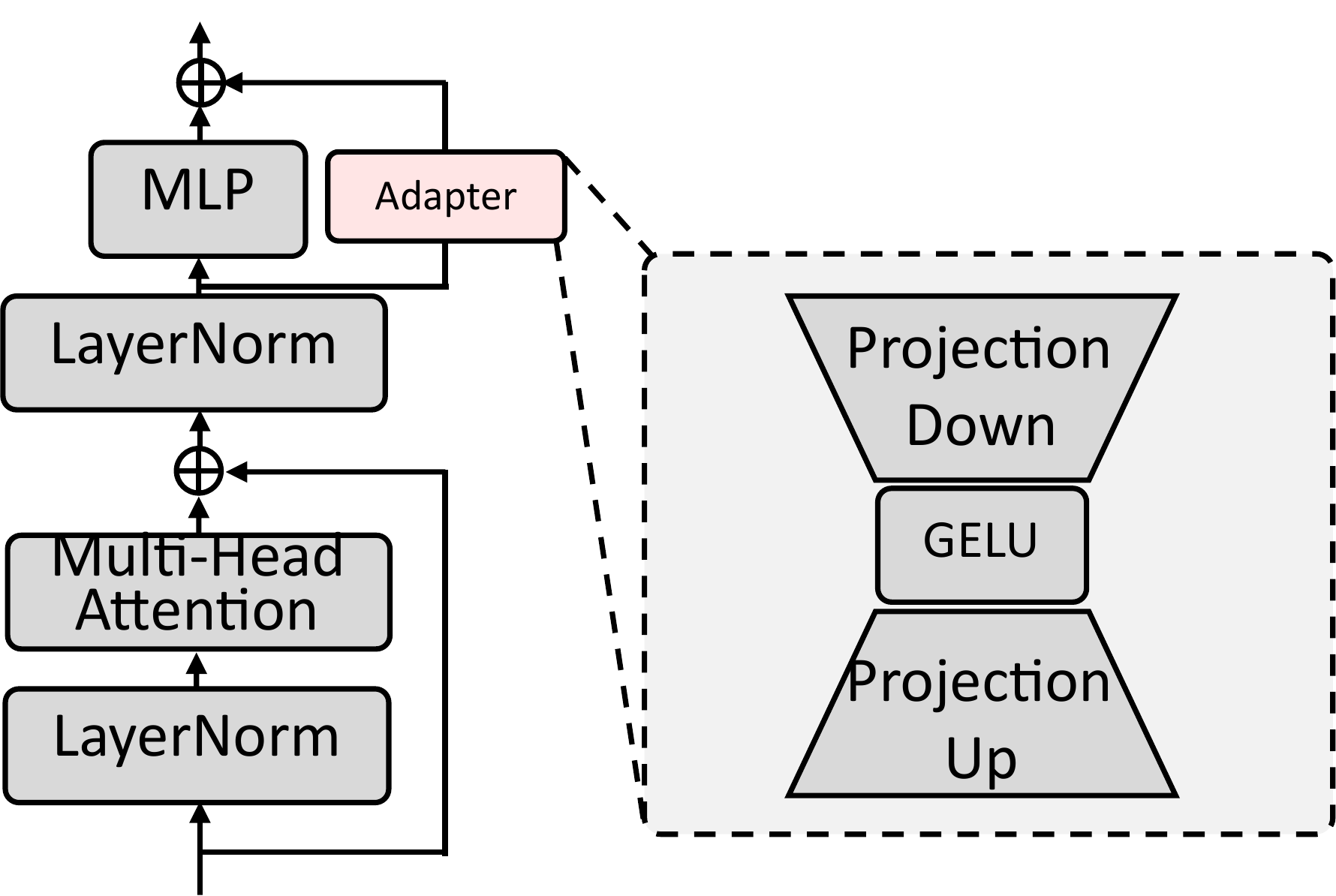}
    \caption{Structure of the Adapter.}
    \label{fig: adapter}
\end{figure}

To address these limitations, we propose the Sparse Adapter. As shown in Fig.~\ref{fig: sparse}, unlike previous approaches~\cite{sam-med2d,samadapter} that insert adapter layers into each encoder block (left side), we insert the adapter only at specific blocks (right side), reducing the number of learnable parameters in the encoder backbone. This design stems from two fundamental considerations: (1) the requirement to prioritize learning generalized curvilinear patterns rather than domain-specific features and (2) the necessity to maintain parameter efficiency while preserving the pretrained encoder's generalization capacity. This will be demonstrated in the ablation study (Table~\ref{tab:adapter_comp}).

Fig.~\ref{fig: adapter} illustrates the adapter's placement after the Multi-head Attention layer. The adapter is implemented as a residual branch placed adjacent to the MLP: it first projects the incoming features to a lower-dimensional space, applies an activation, and then projects them back to the original dimension. The adapter's operation can be mathematically expressed as:

\begin{equation}
  \mathbf{y} = \mathbf{W}_2 \cdot \sigma(\mathbf{W}_1 \cdot \mathbf{x}), 
\end{equation}
where \(\mathbf{x} \in \mathbb{R}^{d}\) and \(\mathbf{y} \in \mathbb{R}^{d}\) are the input and output features, respectively. \(\mathbf{W}_1 \in \mathbb{R}^{r \times d}\) and \(\mathbf{W}_2 \in \mathbb{R}^{d \times r}\) are the down-projection and up-projection matrices, respectively. \(r\) is the bottleneck ratio set to 0.1 in this study. \(\sigma\) denotes the GELU activation function.

\begin{figure}[t]
    \centering
    \subfloat[{\normalfont Inputs}]{
       \includegraphics[width=0.35\linewidth]{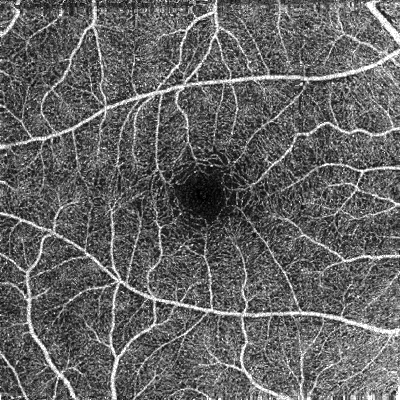}
    }\hfill
    \subfloat[{\normalfont High-Pass Results}]{
       \includegraphics[width=0.35\linewidth]{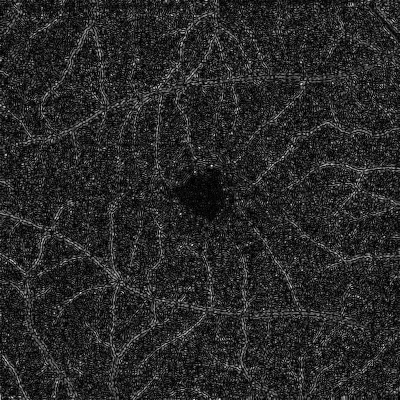}
    }
    \caption{(a) Input image. (b) Result after FFT-based high-pass filtering. The high-pass output preserves and enhances curvilinear structures (e.g., vessels, cracks).}
    \label{highpass}
\end{figure}

\subsubsection{\textbf{Prompt Generation}}\label{pg}
In segmentation tasks involving small objects, such as curvilinear structures, fine structural information tends to attenuate as the number of encoder blocks increases. In order to preserve fine-grained structural details, the PG module, as depicted in Fig.~\ref{UCS}(b), is designed to generate domain-adaptive initial prompts by exploiting frequency-domain representations of the input data. This is effective because curvilinear structures typically differ from the slowly varying background (low‑frequency content), so an FFT-based high‑pass filter suppresses the background while preserving the high‑frequency curve details, which is illustrated in Fig.~\ref{highpass}.

To simplify things, the PG module extracts curve-related cues directly from input images. Specifically, the module transforms the input data into the frequency domain using an FFT. Subsequently, a central rectangular mask is then constructed to suppress low-frequency components, forming a high-pass filter that highlights high-frequency details in the transformed data. The high-pass filter \(H(u,v)\) is defined over frequency coordinates \((u,v)\), with dimensions determined by the input width \(w\), height \(h\), and a tunable parameter \(rate\) (0.25 in this study). The filter is expressed as:

\begin{equation}
\begin{split}
H(u,v) = \begin{cases} 
0, & \text{if } |u - \frac{w}{2}| < \frac{\sqrt{w \cdot h \cdot rate}}{2}, \\
   & \text{and } |v - \frac{h}{2}| < \frac{\sqrt{w \cdot h \cdot rate}}{2}, \\
1, & \text{otherwise} .
\end{cases}
\end{split}
\end{equation}

Subsequently, an inverse FFT is applied to transform the filtered frequency signal back into the spatial domain. As shown in Fig.~\ref{UCS}(b), for the input image $\mathbf{X}$, the prompt embedding $\hat{\mathbf{X}}$ at each encoder block can be represented as:
\begin{equation}\label{eq1}
\hat{\mathbf{X}}_i = MLP_i(L(E(X)) + IFFT(H(FFT(\mathbf{X}))))
\end{equation}
where $MLP_{i}$, $E$, $IFFT$, $H$, and $FFT$ represent the $i$-$th$ Multilayer Perceptron, Patch Embedding layer, Inverse Fast Fourier Transform, high-pass filter, and  Fast Fourier Transform, respectively.

\subsection{Mask Decoder}\label{decoder}
The mask decoder needs to balance global context and local detail for universal curvilinear structure segmentation. We achieve this with a dual-compression architecture using Hierarchical Feature Compression for fine-grained detail and Guidance Feature Compression for automated image-based segmentation. Finally, Mask Embedding integrates these features for refined output.

\subsubsection{\textbf{Hierarchical Feature Compression}}
The purpose of the HFC module is to enrich the decoder by integrating and supplementing features from various blocks of the image encoder. The pipeline of this module is shown in Fig.~\ref{UCS}(c). HFC performs interval sampling by sampler on the encoder layer's outputs. Given $N$(24 in this study) encoder blocks, we sample features at intervals of $\Delta$ (6 in this study). The sample block works as concatenated along the channel dimension to produce a unified feature representation:
\begin{equation}
\label{eq:sample}
    F_{sampled} = Cat(F_{l_{1}}, \ldots, F_{l_{i}}), \quad l_i = i\Delta-1
\end{equation}
where $F_{l_{i}}$ denotes the $l_{i}$-$th$ block features.

We calculated the cosine similarity between the feature map of the first encoder block and each subsequent encoder block, from encoder block \(F_{l_{1}}\) to encoder block \(F_{l_{i}}\). The cosine similarity is computed as:

\begin{equation}
\text{Cosine Similarity} = \frac{F_{l_{1}} \cdot F_{l_{i}}}{\|F_{l_{1}}\| \|F_{l_{i}}\|}, \quad i = 1, 2, \dots, i
\end{equation}

\begin{figure}[t]
    \centering
    \includegraphics[width=0.9\linewidth]{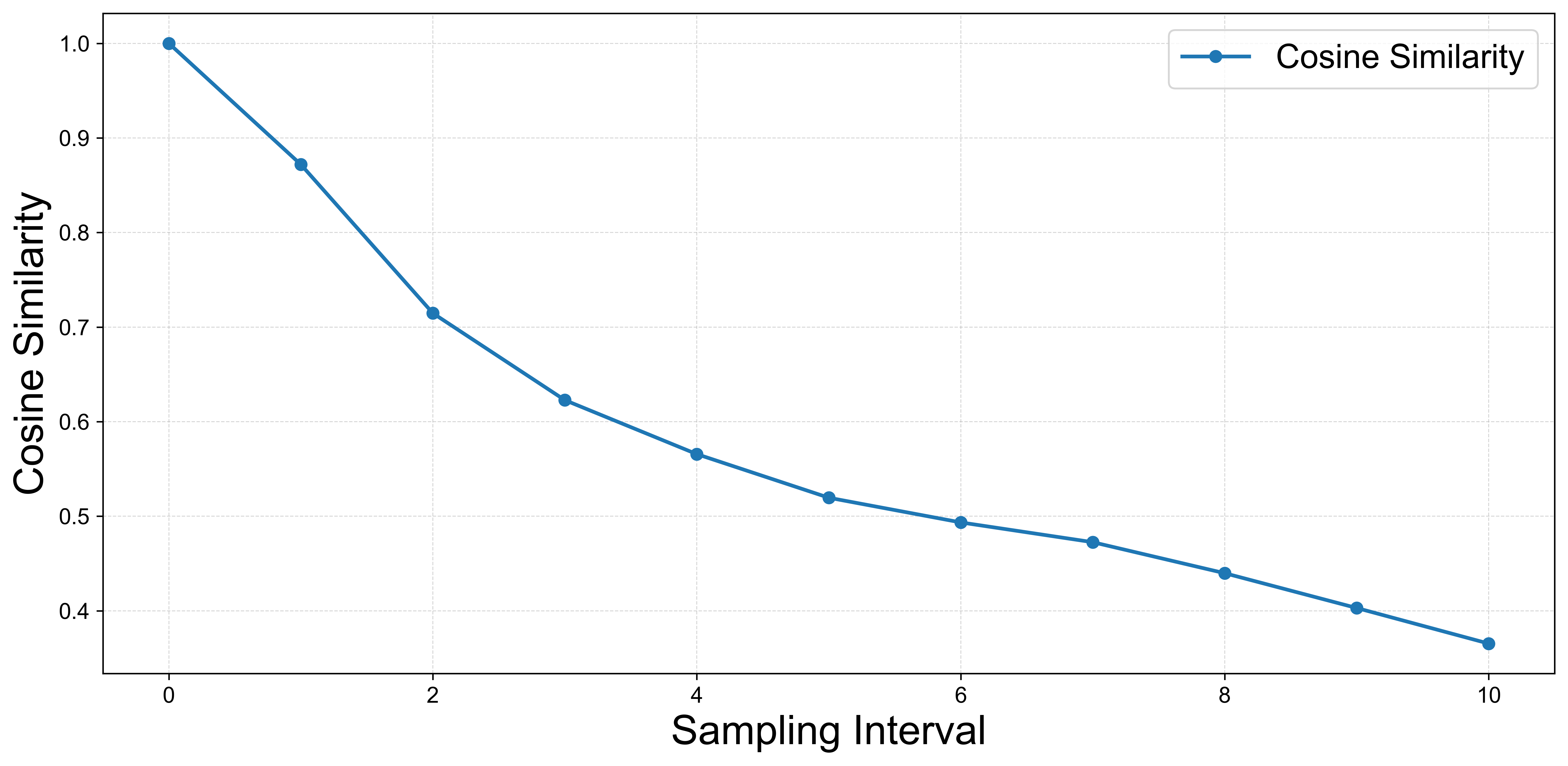}
    \caption{Cosine similarity vs. sampling interval for encoder block feature maps.}
    \label{cos_value}
\end{figure}

As shown in Fig.~\ref{cos_value}, a line graph illustrates the relationship between cosine similarity and the sampling interval for encoder block feature maps. The graph demonstrates a clear downward trend in cosine similarity as the sampling interval increases. Notably, the cosine similarity falls below 0.5 when the sampling interval reaches 6.

\begin{figure}[t]
 \centering
 \subfloat[]{%
     \includegraphics[width=0.49\columnwidth]{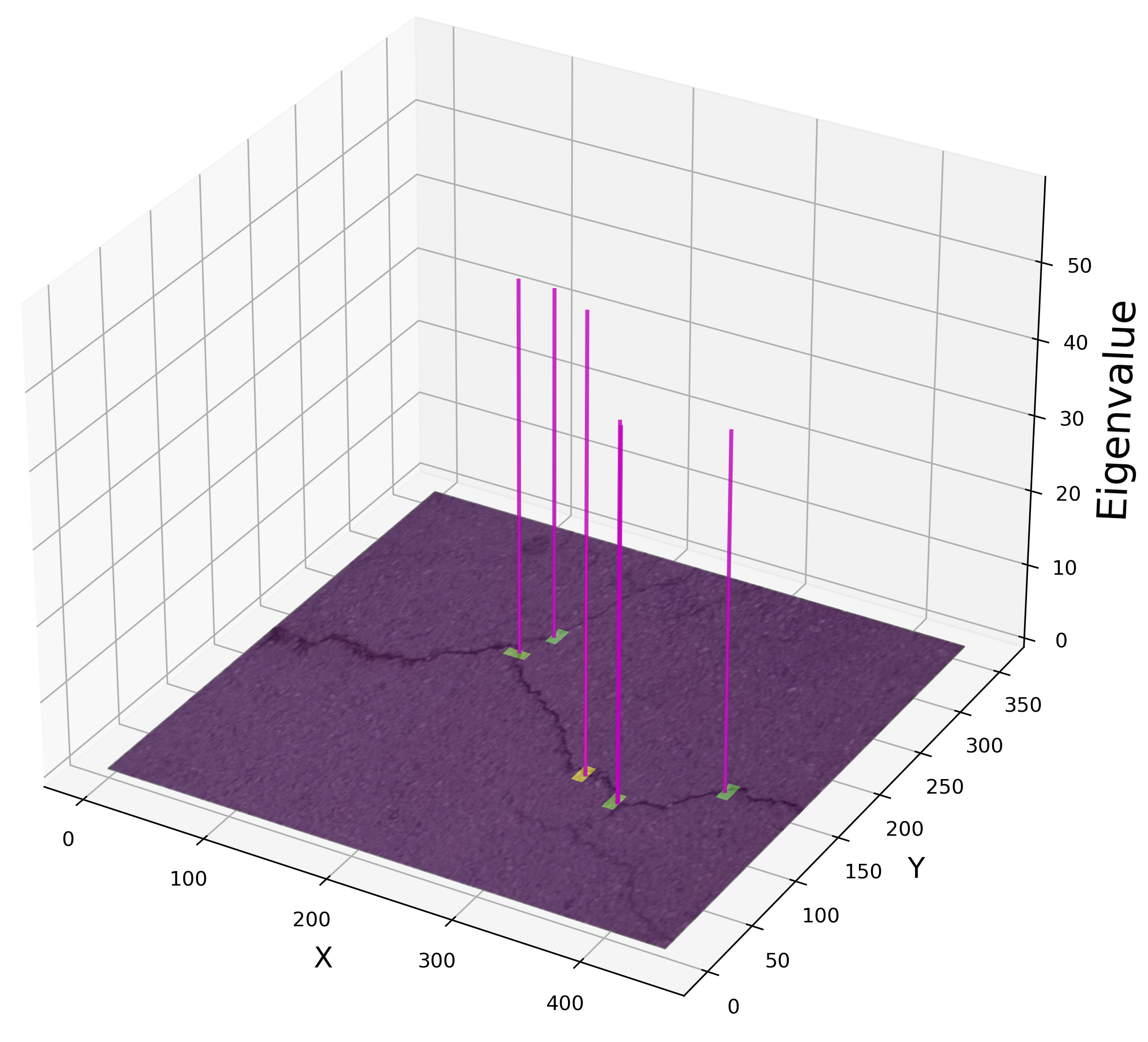}%
 }%
 \hfill
 \subfloat[]{%
     \includegraphics[width=0.49\columnwidth]{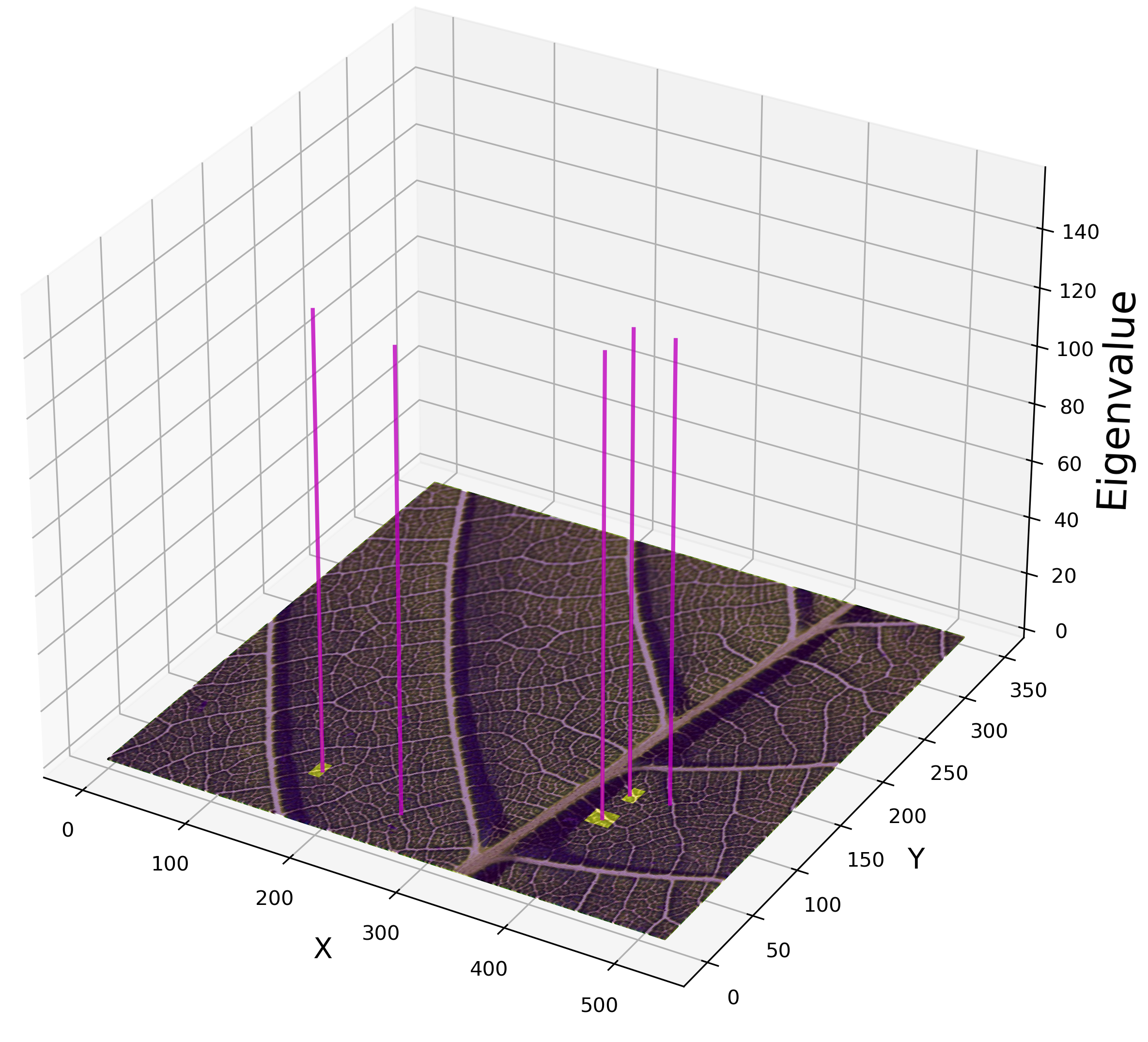}%
 }%
 \caption{Top 5 eigenvalues of the image after PCA.}
 \label{pcashow}
\end{figure}

\subsubsection{\textbf{Guidance Feature Compression}}

Unlike the segmentation of general objects, the elongated and intertwined characteristics of curvilinear structures make SAM's original three interactive prompting methods less effective. Specifically, box-based prompts are not suitable for elongated curves. Point-based and mask-based prompts are cumbersome and imprecise, requiring laborious manual placement. 

Using a 15×15 sliding window, we compute the covariance matrix of pixel intensities across the image and map the top five eigenvalues back to their spatial coordinates. As illustrated in Fig.\ref{pcashow}, regions associated with the highest eigenvalues consistently align with curvilinear structures, validating PCA’s effectiveness in identifying their principal characteristics.

Inspired by PCA and its ability to reduce dimensionality while retaining principal components, we propose the GFC module and integrate it into GFC to automatically capture dominant features. Considering that the dimension of the outputs of the prompt encoder and decoder layer $X \in \mathbb{R}^{5 \times 256}$. PCA is applied to perform dimensionality reduction and guidance feature extraction with the following formula:
\begin{equation}
    \mathbf{X^{\prime}} = ({\mathbf{X} - {\mu}}) \cdot \mathbf{V}_k^{C},
\end{equation}
where ${\mu}$ is the mean of $\mathbf{X}$, $\mathbf{C}$ is the covariance matrix of the $({\mathbf{X} - {\mu}})$, and $\mathbf{V}_k^{C}$ is the projection matrix formed by the $k$ (5 in this study) eigenvectors of $\mathbf{C}$. $\mathbf{X^{\prime}} \in \mathbb{R}^{5 \times 5}$ represents the output of GFC. An ablation study, detailed in Table\ref{tab: ablation}, further demonstrates the impact of this approach on segmentation performance.

\begin{table}[t]
  \centering
  \tiny
  \setlength{\tabcolsep}{4pt}
  \caption{Details of the datasets for experiments.}
  \begin{tabularx}{\columnwidth}{l l >{\raggedright\arraybackslash}X >{\raggedright\arraybackslash}X >{\raggedright\arraybackslash}X >{\raggedright\arraybackslash}X} 
    \toprule
    Split & Distribution & Dataset            & Modality     & Domain      & Target \\
    \midrule
    \multirow{7}{*}{Train} 
      & - & DRIVE~\cite{staal2004ridge}          & Fundus  & Medical     & Vessels \\
      & - & CHASEDB1~\cite{owen2009measuring}    & Fundus  & Medical     & Vessels \\
      & - & CORN-1~\cite{mou2019cs}              & Nerve    & Medical     & Fibers \\
      & - & CRACKTREE200~\cite{zou2012cracktree} & RGB   & Engineering & Cracks \\
      & - & CRACKFOREST~\cite{shi2016automatic}  & RGB   & Engineering & Cracks \\
      & - & XCAD~\cite{SSVS}                     & X-ray   & Medical     & Vessels \\
      & - & ROSSA~\cite{ning2023accurate}        & OCTA    & Medical     & Vessels \\
      \midrule
      \multirow{8}{*}{\makecell{Test \\ (In-House)}} 
      & Seen  & Crack  & RGB & Engineering & Crack \\
      & Unseen  & Branch & RGB & Plant       & Branch \\
      & Unseen  & Floor  & RGB & Engineering & Floor \\
      & Unseen  & Scratch& RGB & Engineering & Scratch \\
      & Unseen  & Soil   & RGB & Natural     & Soil \\
      & Unseen  & Wire   & RGB & Engineering & Wire \\
      & Unseen  & Leaf   & RGB & Plant       & Venation \\
      & Unseen  & Tyre   & RGB & Engineering & Tread line \\
      \midrule
      \multirow{6}{*}{\makecell{Test \\ (Public)}} 
      & unseen & TubeTK~\cite{itktube} & MRA & Medical     & Vessels \\
      & Seen   & OCTA500\_3M~\cite{octa} & OCTA & Medical     & Vessel \\
      & Seen   & OCTA500\_6M~\cite{octa} & OCTA & Medical     & Vessel \\
      & Seen   & CSD~\cite{kaggle_crack_segmentation_dataset} & RGB & Engineering & Crack \\
      & Seen   & DCA1~\cite{DCA1} & X-ray & Medical     & Vessels \\
      & Seen   & STARE~\cite{stare} & Fundus & Medical     & Vessel \\
      & Seen   & FIVES~\cite{jin2022fives} & Fundus & Medical     & Vessel \\
    \bottomrule
  \end{tabularx}
  \label{tab:datasets}
\end{table}

\subsection{Loss Formulation}\label{loss}
A total loss function is designed by combining Focal Loss~\cite{focalloss}, Dice Loss~\cite{diceloss}, and Mask IoU Loss~\cite{maskiou} for segmentation of curvilinear structures. It's defined as:
\begin{equation}
\mathcal{L}_{\text{total}} = \alpha \cdot \mathcal{L}_{\text{focal}} + \beta\ \cdot \mathcal{L}_{\text{dice}} + \gamma \cdot \mathcal{L}_{\text{IoU}},
\end{equation}
where \(\alpha\), \(\beta\), \(\gamma\) are the weight coefficients for Focal Loss, Dice Loss, and Mask IoU Loss, respectively.

The calculation formulas for these loss functions are as follows:
\begin{equation}
\left\{
\begin{aligned}
\mathcal{L}_{\text{focal}} &= - \frac{1}{N} \sum_{i=1}^{N} 
\begin{cases}
(1 - p_i)^\zeta  \log(p_i), & \text{if } y_i = 1 \\
p_i^\zeta  \log(1 - p_i), & \text{if } y_i = 0
\end{cases} \\
\mathcal{L}_{\text{dice}} &= 1 - \frac{2 \sum_{i=1}^{N} p_i y_i + \epsilon}{\sum_{i=1}^{N} p_i^{2} + \sum_{i=1}^{N} y_i^{2} + \epsilon} \\
\mathcal{L}_{\text{IoU}} &= 1 - \frac{\sum_{i=1}^{N} p_i y_i}{\sum_{i=1}^{N} p_i + \sum_{i=1}^{N} y_i - \sum_{i=1}^{N} p_i y_i + \epsilon}
\end{aligned}
\right.
\end{equation}
where \(\zeta\) and \(\epsilon\) are the focusing parameter and smoothing term with default values. Typical values for \(\alpha\), \(\beta\), and \(\gamma\) are determined empirically. For instance, we empirically set $\alpha = 20.0$, $\beta = 1.0$, and $\gamma = 1.0$, which showed good performance in our experiments.

\section{Experiment and Evaluation}
\subsection{Experimental Setup}
\subsubsection{Train Dataset}
We curated a comprehensive training set by gathering existing datasets for curvilinear segmentation: DRIVE\cite{staal2004ridge}, CHASEDB1\cite{owen2009measuring}, CORN-1\cite{mou2019cs}, CRACKTREE200\cite{zou2012cracktree}, CRACKFOREST\cite{shi2016automatic}, XCAD\cite{SSVS}, and ROSSA\cite{ning2023accurate}. These datasets, referred to as the Train Dataset, consist of 3,244 annotated image-mask pairs, providing a solid foundation for training.

\subsubsection{Test Dataset}
We created a novel {\textbf{In-House Test Dataset}} to evaluate open-set segmentation capabilities across diverse and challenging scenarios. This dataset is publicly available at IEEE DataPort (doi: https://dx.doi.org/10.21227/1n2q-6770). This dataset comprises images of curvilinear structures captured across eight distinct natural scene categories: branch, crack, floor, scratch, soil, wire, leaf, and tyre. Representative images from these categories are shown in Fig.~\ref{demoresult}(a). Each image in the dataset was annotated with pixel-level accuracy through manual labeling to ensure high-quality ground truth masks. 

Additionally, we collected seven additional benchmark datasets for curvilinear segmentation to evaluate our model, collectively referred to as the {\textbf{Public Test Dataset}}: OCTA500\_3M~\cite{octa}, OCTA500\_6M\footnote{The images in OCTA500\_3M and OCTA500\_6M share some similarities with those in the ROSSA~\cite{ning2023accurate} from the Train Dataset.}~\cite{octa}, CSD\footnote{The Crack Segmentation Dataset (CSD) is derived from~\cite{kaggle_crack_segmentation_dataset}, which was created by merging multiple crack segmentation datasets~\cite{shi2016automatic, yang2019feature, eisenbach2017how, amhaz2016automatic, DeeperCrack, zou2012cracktree} and included additional non-crack images. The original test folder contains 1,695 images. To avoid overlap with the Train Dataset, we excluded 31 images from CRACKTREE200~\cite{zou2012cracktree}, 18 images from CRACKFOREST~\cite{shi2016automatic}, and 212 nocrack images. The final CSD used in our experiments consisted of 1,434 unique image-mask pairs.}, DCA1~\cite{DCA1}, TubeTk~\cite{itktube}, STARE~\cite{stare}, and FIVES~\cite{jin2022fives}.
The combination of the In-House Test Dataset and the Public Test Dataset forms a comprehensive benchmark, enabling the evaluation of segmentation performance on diverse and unseen domains. Here, "seen" and "unseen" refer to imaging modalities or segmentation targets that are, respectively, present or absent in the training set. The details of the training and test splits, included datasets, imaging modalities, and the corresponding segmentation targets are summarized in Table~\ref{tab:datasets}.

\subsubsection{Implementation Details}\label{sec: train}
UCS and the comparison models were either fine-tuned or trained on the Train Dataset until convergence to ensure a fair evaluation of their generalization capabilities. 

UCS was implemented using PyTorch 2.4.0 and CUDA 11.8. To augment the data, we employed horizontal flipping and random cropping techniques. The model was trained for 10 epochs with an initial learning rate was set to \(1\times10^{-4}\), and the model was trained using the Adam optimizer with a batch size of 1. The binarization threshold was set to 0.5. The comparison models were trained or fine-tuned with default hyperparameters from their open-source implementations. All models, including UCS and the comparison models, were trained on an NVIDIA GeForce RTX 3090 GPU.

\subsubsection{Evaluation metrics}
For pixel-based evaluation, we employ the F1-score, Precision, and Recall metrics, which are widely adopted in existing semantic segmentation methods~\cite{bloodsegment, DeepCD, yu, FR-UNet, SFIAN, MDAU-net}. 

\setlength{\tabcolsep}{0.5pt} %
\begin{figure*}[t]
\centering
\begin{tabularx}{\textwidth}{
    >{\raggedleft\arraybackslash}m{0.3cm} %
    *{7}{>{\centering\arraybackslash}m{0.138\textwidth}} 
}
\rotatebox[origin=c]{90}{\parbox[c][0.3cm]{1.5cm}{\centering \textbf{FIVES}}} & 
\includegraphics[width=\linewidth, height=3cm, keepaspectratio]{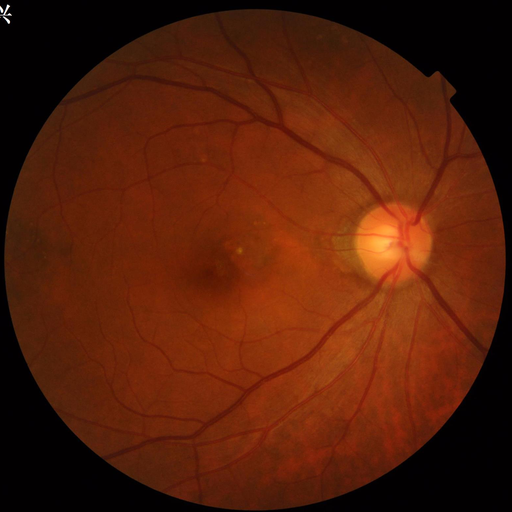} & 
\includegraphics[width=\linewidth, height=3cm, keepaspectratio]{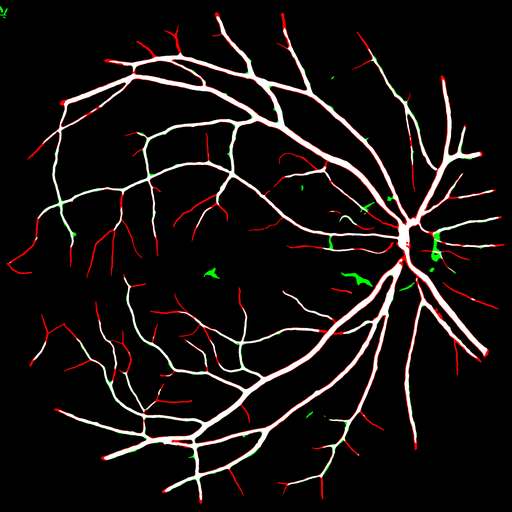} &
\includegraphics[width=\linewidth, height=3cm, keepaspectratio]{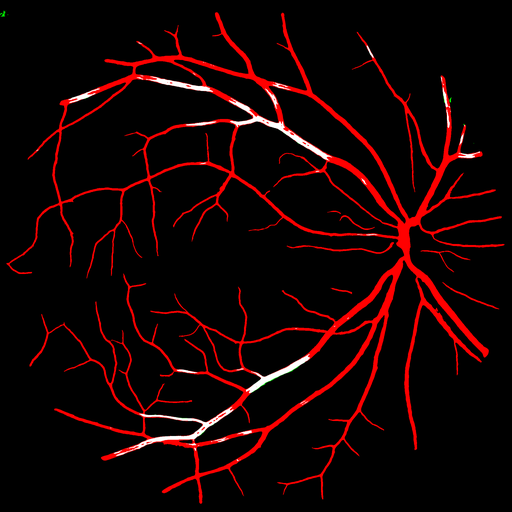} & 
\includegraphics[width=\linewidth, height=3cm, keepaspectratio]{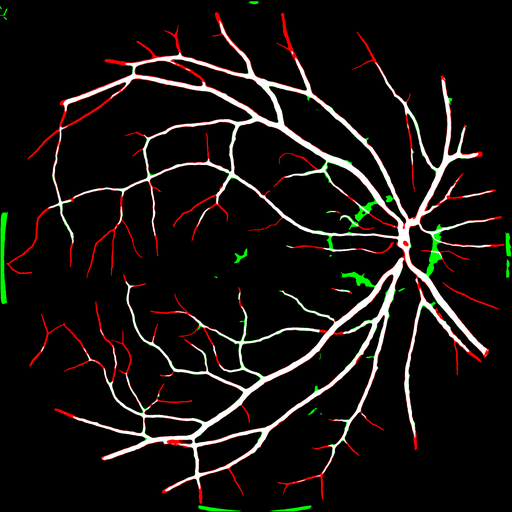} & 
\includegraphics[width=\linewidth, height=3cm, keepaspectratio]{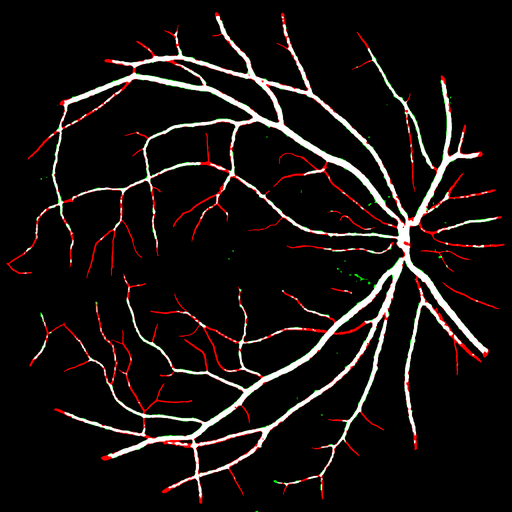} & 
\includegraphics[width=\linewidth, height=3cm, keepaspectratio]{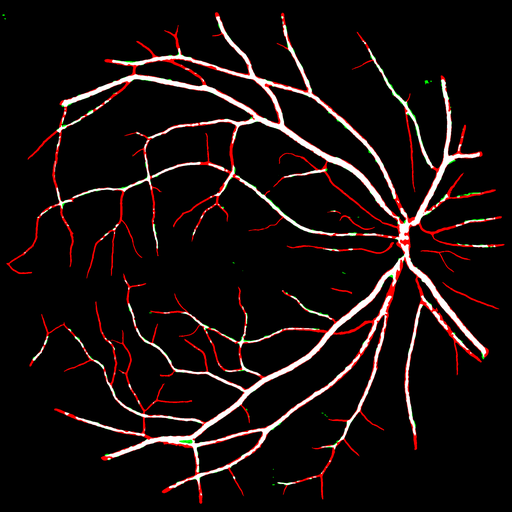} & 
\includegraphics[width=\linewidth, height=3cm, keepaspectratio]{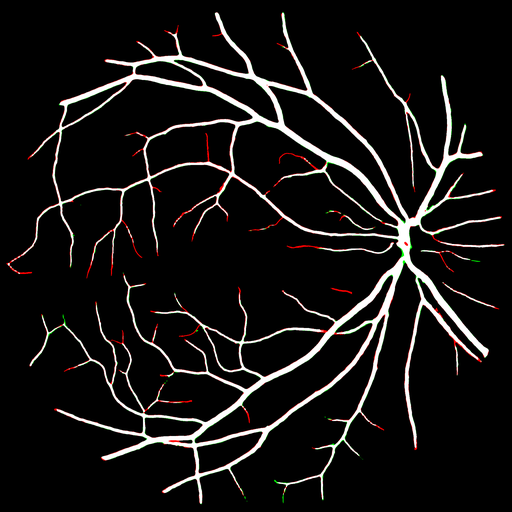} \\[-2pt]

\rotatebox[origin=c]{90}{\parbox[c][0.3cm]{1.5cm}{\centering \textbf{OCTA\_3M}}} & 
\includegraphics[width=\linewidth, height=3cm, keepaspectratio]{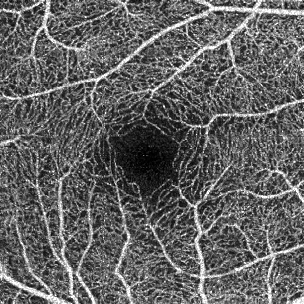} & 
\includegraphics[width=\linewidth, height=3cm, keepaspectratio]{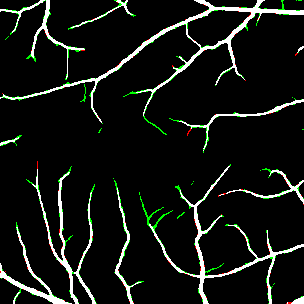} & 
\includegraphics[width=\linewidth, height=3cm, keepaspectratio]{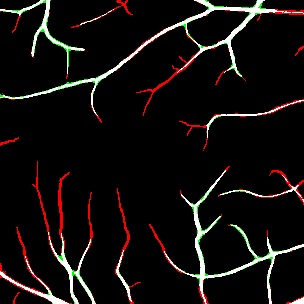} & 
\includegraphics[width=\linewidth, height=3cm, keepaspectratio]{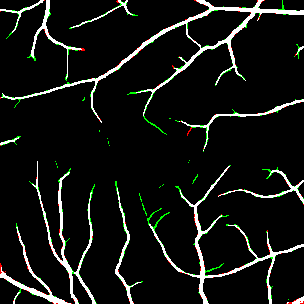} & 
\includegraphics[width=\linewidth, height=3cm, keepaspectratio]{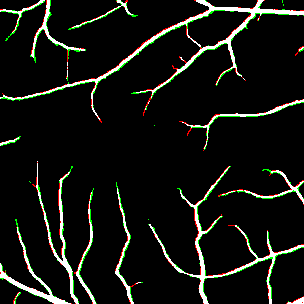} & 
\includegraphics[width=\linewidth, height=3cm, keepaspectratio]{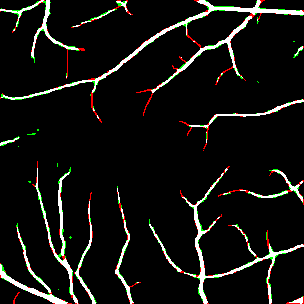} & 
\includegraphics[width=\linewidth, height=3cm, keepaspectratio]{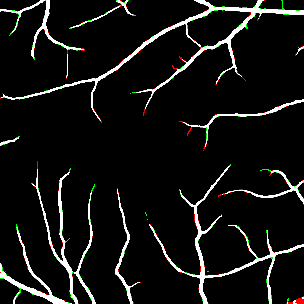} \\[-2pt]

\rotatebox[origin=c]{90}{\parbox[c][0.3cm]{1.5cm}{\centering \textbf{TubeTK}}} & 
\includegraphics[width=\linewidth, height=3cm, keepaspectratio]{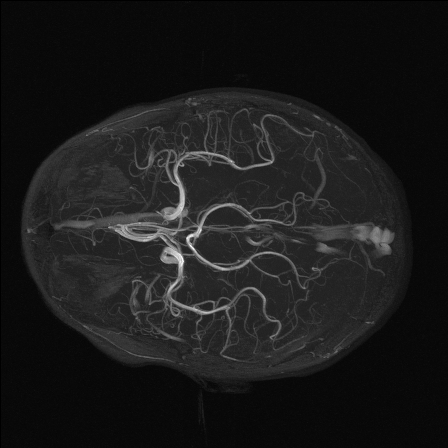} & 
\includegraphics[width=\linewidth, height=3cm, keepaspectratio]{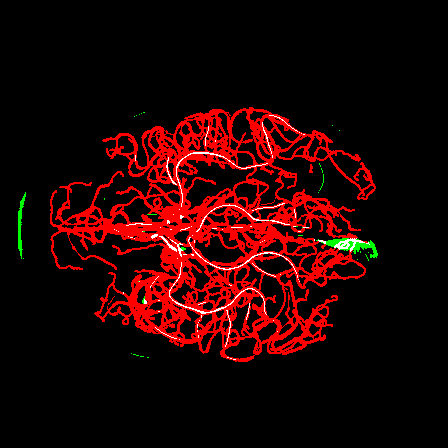} & 
\includegraphics[width=\linewidth, height=3cm, keepaspectratio]{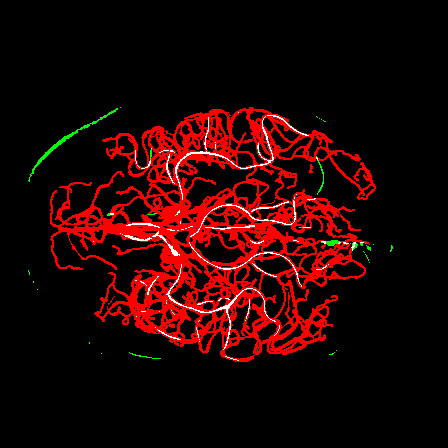} & 
\includegraphics[width=\linewidth, height=3cm, keepaspectratio]{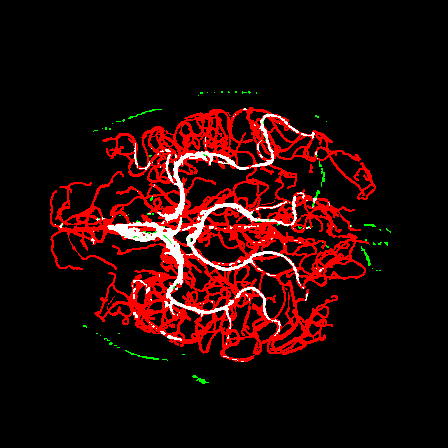} & 
\includegraphics[width=\linewidth, height=3cm, keepaspectratio]{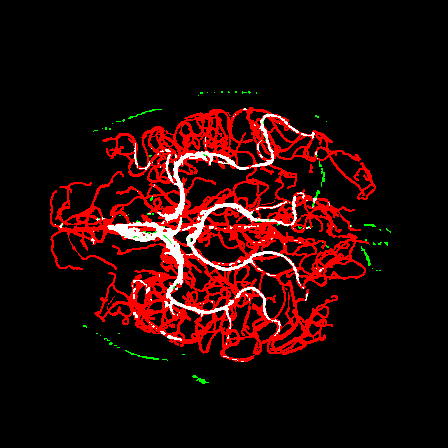} & 
\includegraphics[width=\linewidth, height=3cm, keepaspectratio]{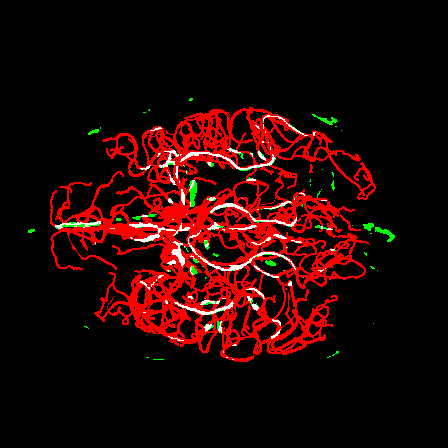} & 
\includegraphics[width=\linewidth, height=3cm, keepaspectratio]{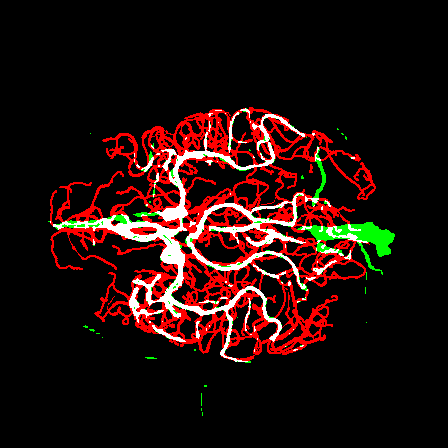} \\[-2pt]

\rotatebox[origin=c]{90}{\raisebox{3.0ex}{\parbox[l][4cm]{1.5cm}{\centering \textbf{LEAF$^{\ast}$}}}} &
\includegraphics[width=\linewidth, height=3cm, keepaspectratio]{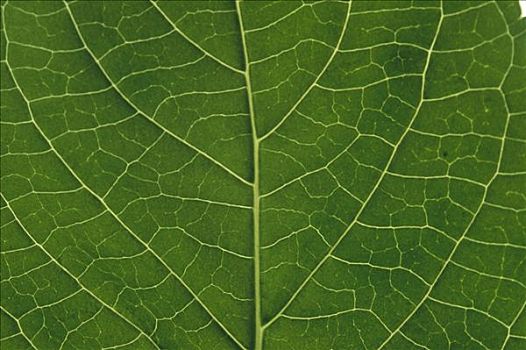} &  
\includegraphics[width=\linewidth, height=3cm, keepaspectratio]{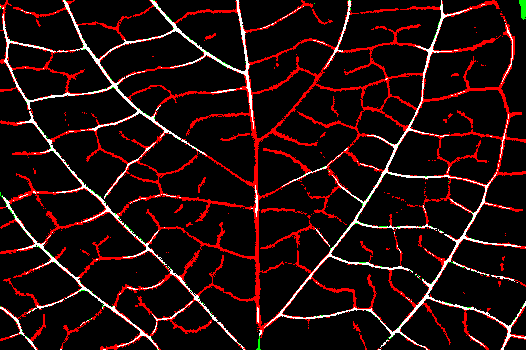} &
\includegraphics[width=\linewidth, height=3cm, keepaspectratio]{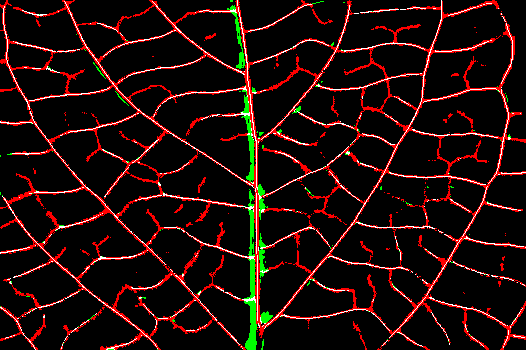} & 
\includegraphics[width=\linewidth, height=3cm, keepaspectratio]{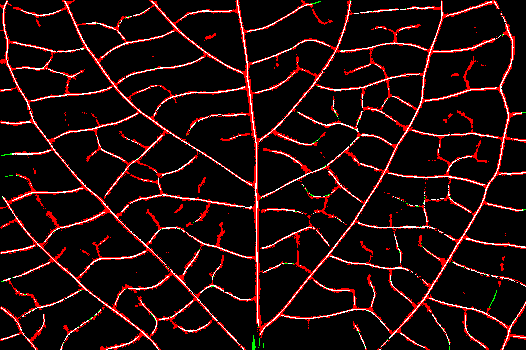} &
\includegraphics[width=\linewidth, height=3cm, keepaspectratio]{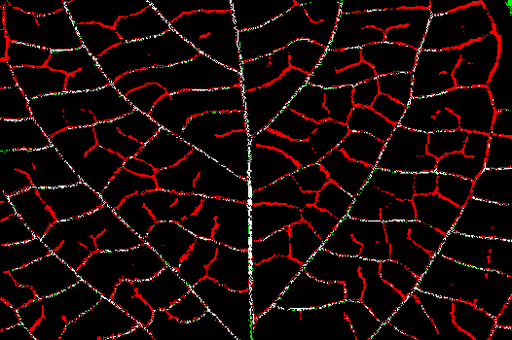} & 
\includegraphics[width=\linewidth, height=3cm, keepaspectratio]{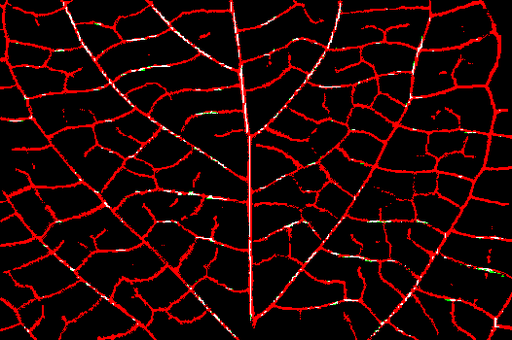} & 
\includegraphics[width=\linewidth, height=3cm, keepaspectratio]{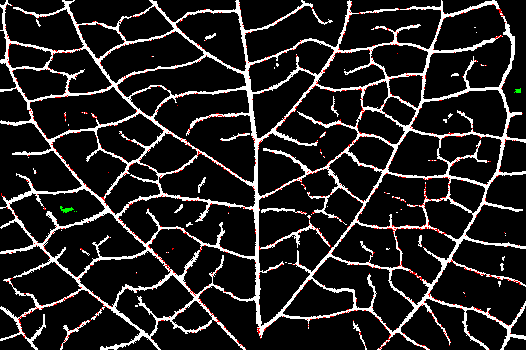} \\[-2pt]

\rotatebox[origin=c]{90}{\raisebox{5.0ex}{\parbox[l][4cm]{1.5cm}{\centering \textbf{CRACK$^{\ast}$}}}} &
\includegraphics[width=\linewidth, height=3cm, keepaspectratio]{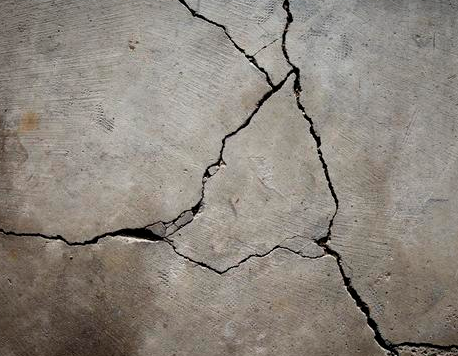} &  
\includegraphics[width=\linewidth, height=3cm, keepaspectratio]{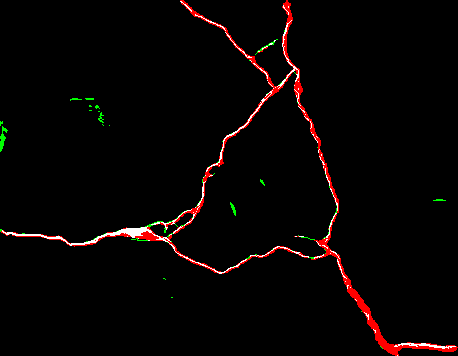} & 
\includegraphics[width=\linewidth, height=3cm, keepaspectratio]{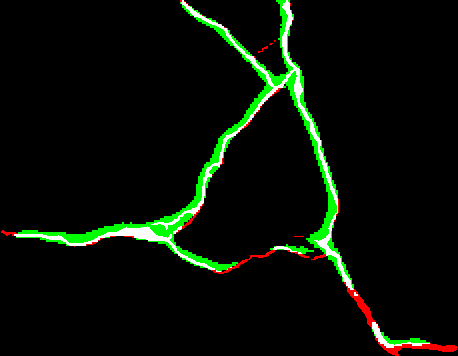} & 
\includegraphics[width=\linewidth, height=3cm, keepaspectratio]{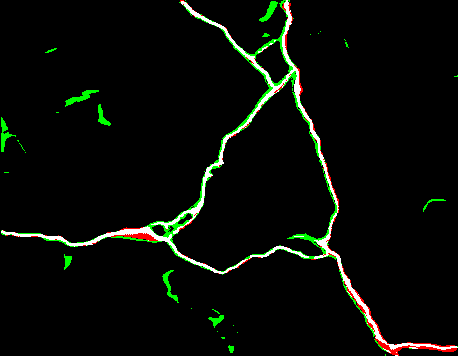} &
\includegraphics[width=\linewidth, height=3cm, keepaspectratio]{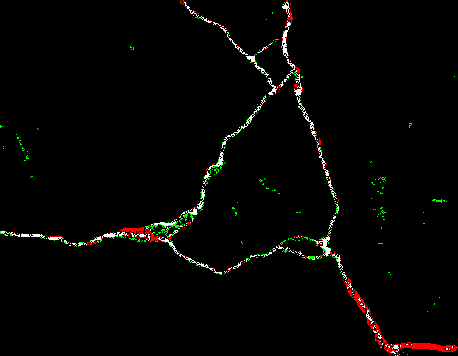} & \includegraphics[width=\linewidth, height=3cm, keepaspectratio]{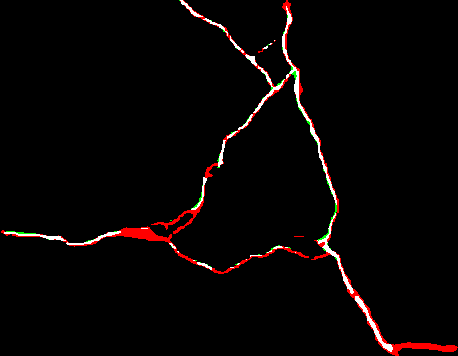} & 
\includegraphics[width=\linewidth, height=3cm, keepaspectratio]{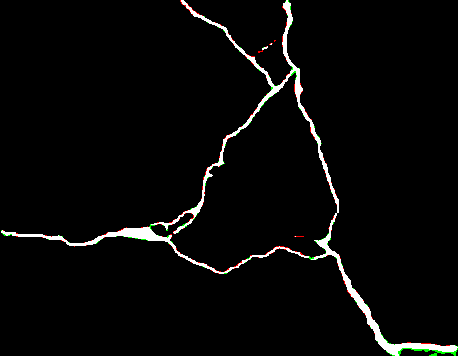} \\[-2pt]

\rotatebox[origin=c]{90}{\raisebox{3.0ex}{\parbox[l][4cm]{1.5cm}{\centering \textbf{SCRATCH$^{\ast}$}}}} &
\includegraphics[width=\linewidth, height=3cm, keepaspectratio]{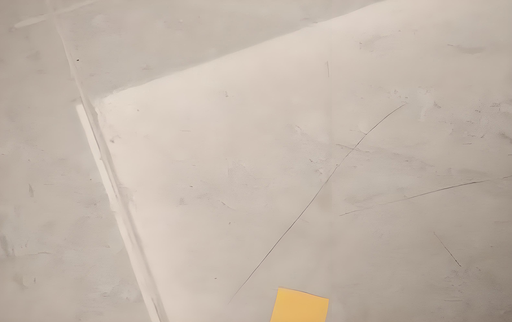} & 
\includegraphics[width=\linewidth, height=3cm, keepaspectratio]{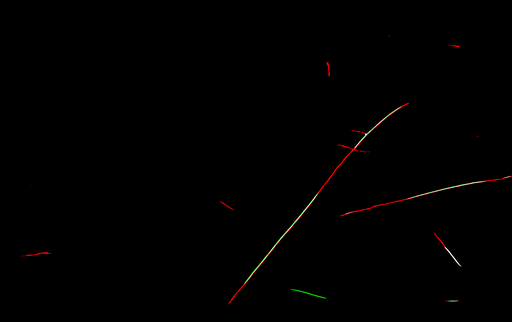} &
\includegraphics[width=\linewidth, height=3cm, keepaspectratio]{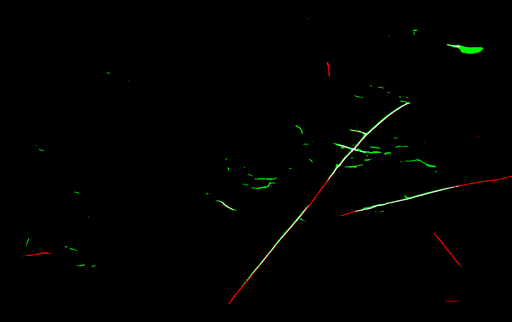} & 
\includegraphics[width=\linewidth, height=3cm, keepaspectratio]{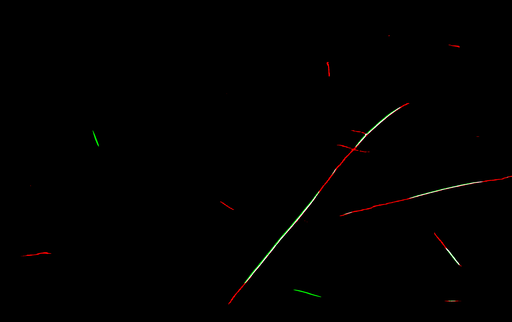} & 
\includegraphics[width=\linewidth, height=3cm, keepaspectratio]{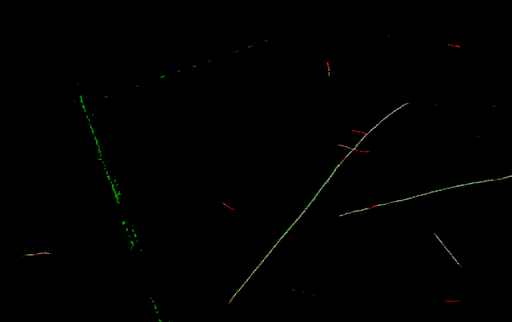} & 
\includegraphics[width=\linewidth, height=3cm, keepaspectratio]{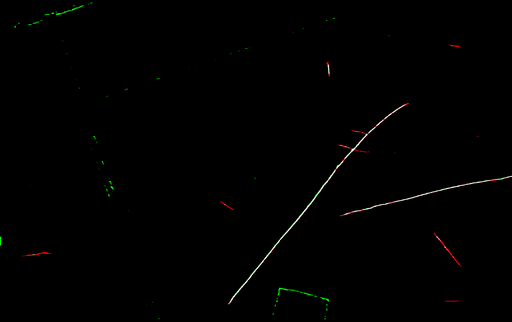} & 
\includegraphics[width=\linewidth, height=3cm, keepaspectratio]{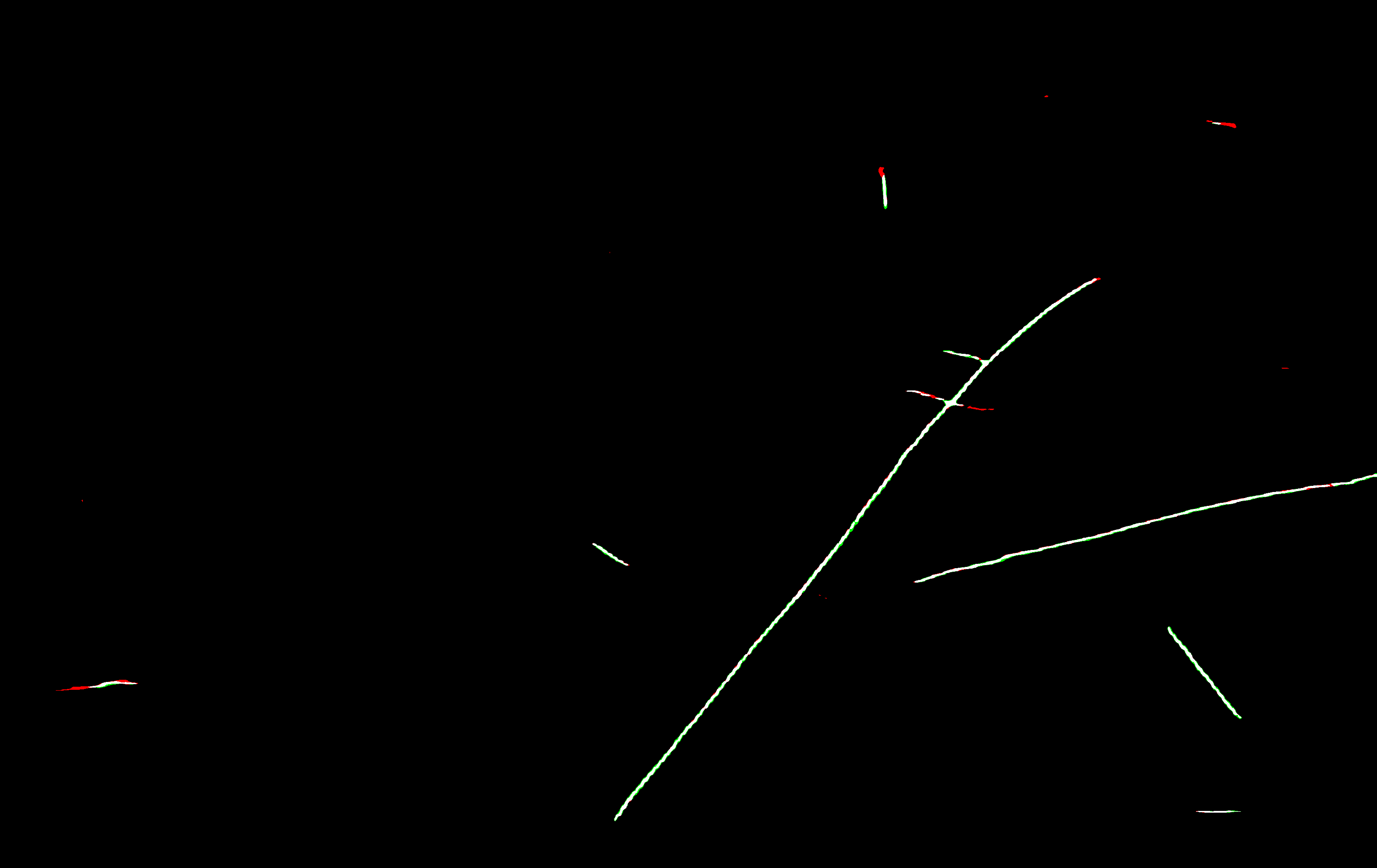} \\[-2pt]

\rotatebox[origin=c]{90}{\raisebox{3.0ex}{\parbox[c][0.3cm]{1.5cm}{\centering \textbf{SOIL$^{\ast}$}}}} & 
\includegraphics[width=\linewidth, height=3cm, keepaspectratio]{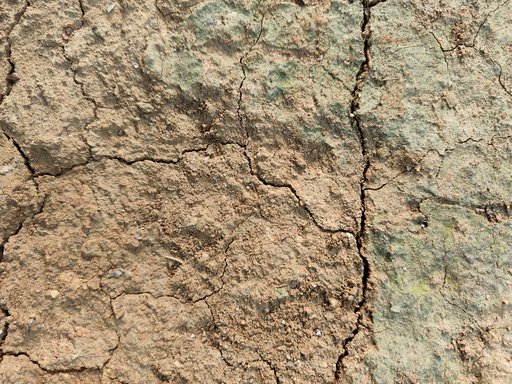} & 
\includegraphics[width=\linewidth, height=3cm, keepaspectratio]{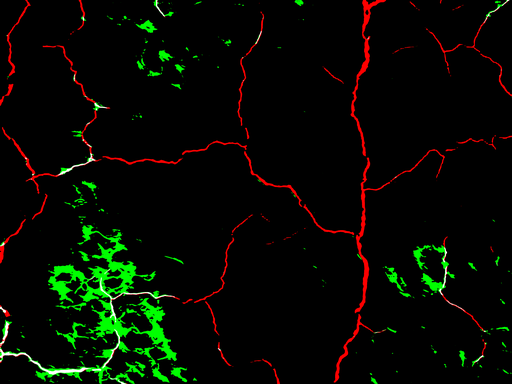} &
\includegraphics[width=\linewidth, height=3cm, keepaspectratio]{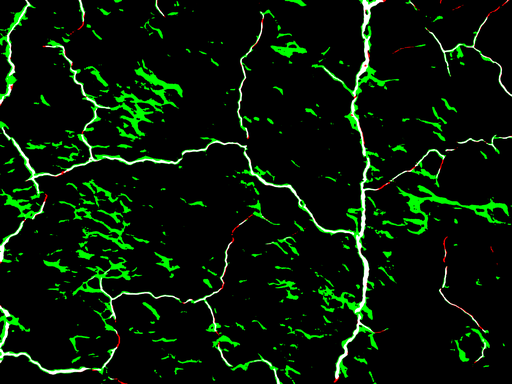} & 
\includegraphics[width=\linewidth, height=3cm, keepaspectratio]{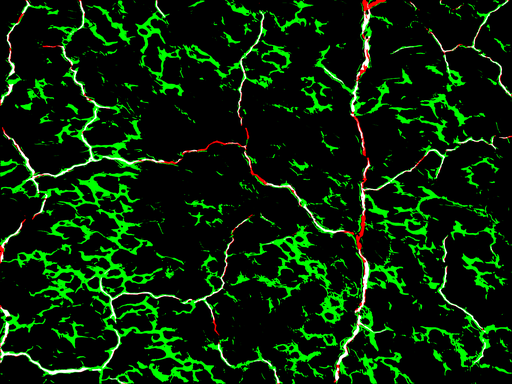} &
\includegraphics[width=\linewidth, height=3cm, keepaspectratio]{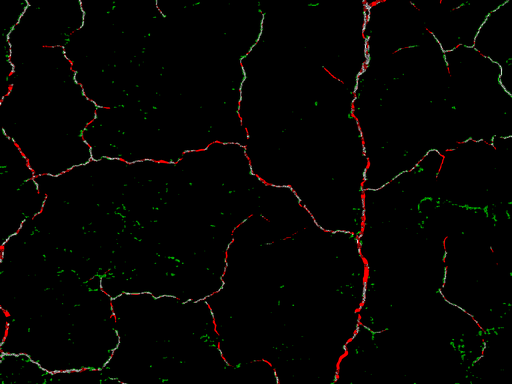} & 
\includegraphics[width=\linewidth, height=3cm, keepaspectratio]{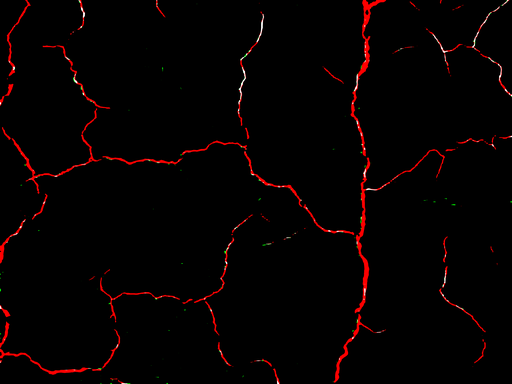} & 
\includegraphics[width=\linewidth, height=3cm, keepaspectratio]{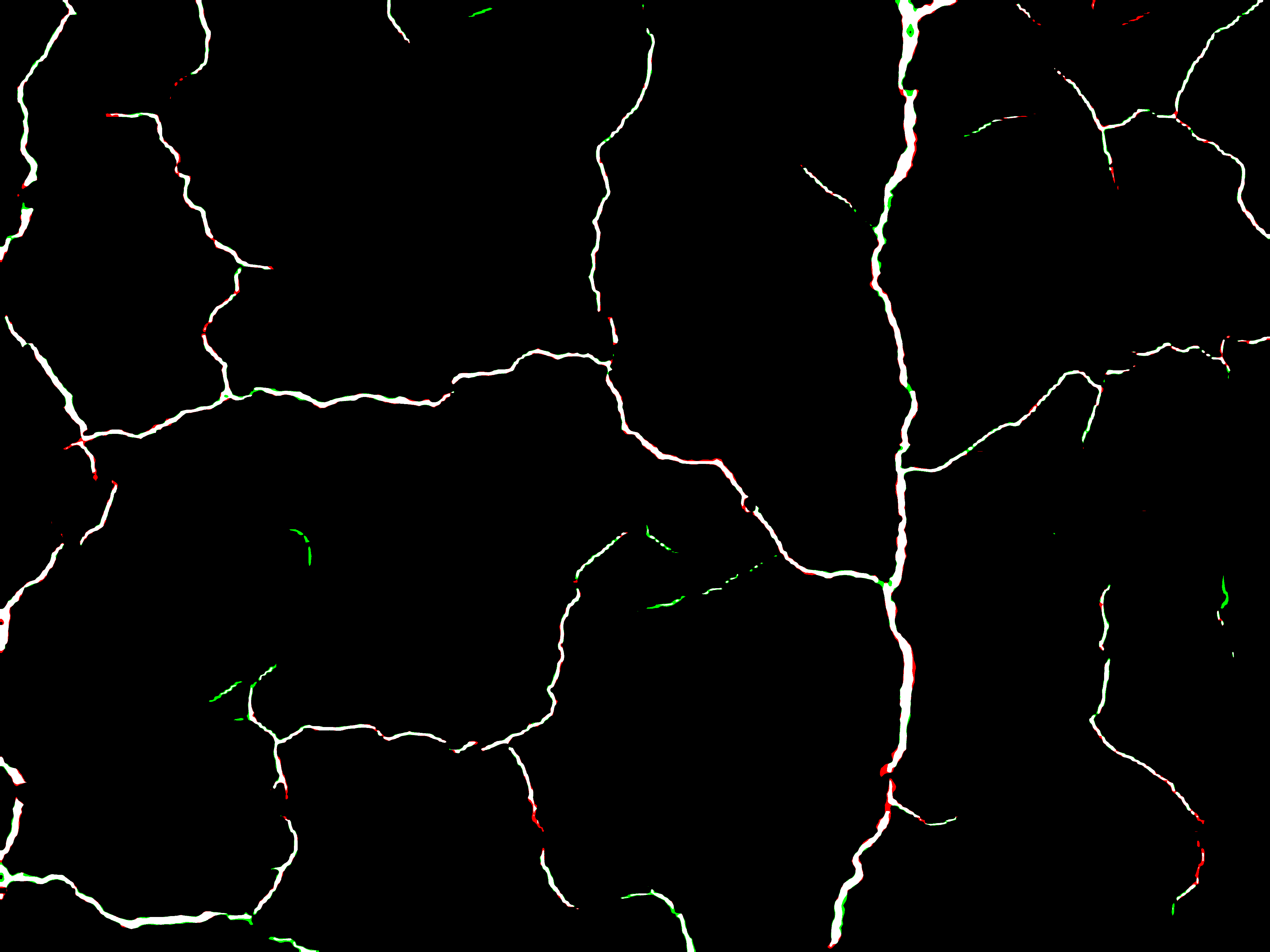} \\[-8pt]
\multicolumn{1}{c}{} & \multirow{2}{*}{\textbf{Input}} & \multirow{2}{*}{\textbf{UNet++~\cite{unet++}}} & \multirow{2}{*}{\textbf{CSSR~\cite{CSSR}}} & \multirow{2}{*}{\textbf{FR-UNet~\cite{FR-UNet}}} & \multirow{2}{*}{\textbf{HQ-SAM~\cite{SAMinHQ}}} & \multirow{2}{*}{\textbf{Rein~\cite{Regin}}} & \multirow{2}{*}{\textbf{Ours}}\\[5pt]
\end{tabularx}
\caption{Qualitative comparison of UCS with other methods across datasets from diverse domains. The left text identifies the image dataset, while the bottom text denotes the segmentation model used. The results are presented as pixel-level comparisons with the ground truth, where black, white, red, and green pixels represent true negatives (TN), true positives (TP), false negatives (FN), and false positives (FP), respectively.}
\label{fig: visualization compare}
\end{figure*}

\subsection{One Model Weight for All}\label{one weight}
We compared with nine different representative models, providing a comprehensive analysis of performance across different architectures. Specifically, we benchmarked our method against general CNN-based frameworks such as UNet-vgg16~\cite{2015unet}, UNet-resnet101~\cite{2015unet}, UNet++~\cite{unet++}, CSNet~\cite{csnet}, CS$^2$Net~\cite{cs2net}, CSSR~\cite{CSSR}, FR-UNet~\cite{FR-UNet}, BCU-Net~\cite{BCU}, as well as larger Transformer-based architectures, including SAM-Med2d~\cite{sam-med2d}, Rein~\cite{Regin}, and HQ-SAM~\cite{SAMinHQ}. Moreover, since foundation models (SAM-Med2d~\cite{sam-med2d}, Rein~\cite{Regin}, and HQ-SAM~\cite{SAMinHQ}) are designed for general purposes, we evaluate both their original pretrained weights and their fine-tuned versions on the Train Dataset for a comprehensive comparison.

\begin{table*}[!t]
    \setlength\tabcolsep{3pt} 
    \centering
    \caption{Comparison results on the {\textbf{In-House Test Dataset}}. Here, $\ddagger$ and $\dagger$ denote test results for models trained and fine-tuned on the Train Dataset, respectively. The best and second-best performances are bolded and underlined, respectively.}
    \label{tab: our data}
    \resizebox{\textwidth}{!}{%
        \begin{tabular}{l|l|ccc|ccc|ccc|ccc|ccc|ccc|ccc|ccc}
            \toprule
            \multirow{2}{*}{Method} & \multirow{2}{*}{Year}
                & \multicolumn{3}{c|}{BRANCH} & \multicolumn{3}{c|}{CRACK} & \multicolumn{3}{c|}{FLOOR}
                & \multicolumn{3}{c|}{SCRATCH} & \multicolumn{3}{c|}{SOIL} & \multicolumn{3}{c|}{WIRE}
                & \multicolumn{3}{c|}{LEAF}   & \multicolumn{3}{c}{TYRE} \\
            \cmidrule(lr){3-5} \cmidrule(lr){6-8} \cmidrule(lr){9-11} \cmidrule(lr){12-14}
            \cmidrule(lr){15-17} \cmidrule(lr){18-20} \cmidrule(lr){21-23} \cmidrule(lr){24-26}
            & & F1 $\uparrow$ & Pre $\uparrow$ & Rec $\uparrow$ 
              & F1 $\uparrow$ & Pre $\uparrow$ & Rec $\uparrow$
              & F1 $\uparrow$ & Pre $\uparrow$ & Rec $\uparrow$
              & F1 $\uparrow$ & Pre $\uparrow$ & Rec $\uparrow$
              & F1 $\uparrow$ & Pre $\uparrow$ & Rec $\uparrow$
              & F1 $\uparrow$ & Pre $\uparrow$ & Rec $\uparrow$
              & F1 $\uparrow$ & Pre $\uparrow$ & Rec $\uparrow$
              & F1 $\uparrow$ & Pre $\uparrow$ & Rec $\uparrow$ \\
            \midrule
            UNet-vgg16~\cite{2015unet} $\ddagger$ 
              & 2015
              & 25.41 & 38.31 & 19.01
              & 22.19 & 47.70 & 14.46
              & 50.21 & 54.88 & 46.28
              & 32.38 & 51.42 & 23.64
              & 13.64 & 12.87 & 14.52
              & 28.72 & 39.03 & 22.73
              & 29.74 & 48.61 & 21.43
              & 44.04 & 33.44 & 64.48 \\
            UNet-resnet101~\cite{2015unet} $\ddagger$ 
              & 2015
              & 16.43 & 30.43 & 11.25
              & 34.96 & \underline{68.33} & 23.49
              & 18.68 & \underline{63.31} & 10.96
              & 20.51 & \underline{60.66} & 12.34
              & 10.64 & 44.04 & 6.05
              & 30.61 & 56.82 & 20.95
              & 22.03 & \textbf{72.97} & 12.97
              & 29.45 & 39.60 & 23.44 \\
            UNet++~\cite{unet++} $\ddagger$ 
              & 2018
              & 36.69 & 41.35 & 25.00
              & 41.15 & 46.21 & 37.10
              & 52.50 & 45.70 & 61.69
              & 34.40 & 49.22 & 26.44
              & 18.31 & 10.55 & 69.38
              & 31.42 & 40.01 & 25.86
              & 21.01 & 26.81 & \underline{77.32}
              & \underline{46.02} & 32.76 & \underline{77.32} \\
            CSNet~\cite{csnet} $\ddagger$ 
              & 2019
              & 31.16 & \underline{54.61} & 29.20
              & 25.80 & 54.12 & 24.23
              & 46.88 & 49.44 & 54.83
              & 35.60 & 44.81 & 35.40
              & 19.10 & 15.90 & 31.66
              & 29.33 & 35.09 & 32.63
              & 11.91 & 14.92 & 12.19
              & 28.72 & 19.99 & 70.02 \\
            CS$^2$Net~\cite{cs2net} $\ddagger$ 
              & 2020
              & 35.62 & 52.87 & 34.77
              & 24.42 & 49.38 & 22.87
              & 44.14 & 47.97 & 52.63
              & 35.42 & 44.17 & 36.21
              & 13.53 & 9.91  & 30.72
              & 27.12 & 30.38 & 33.68
              & 13.65 & 15.29 & 14.64
              & 25.58 & 17.53 & 65.10 \\
            CSSR~\cite{CSSR} $\ddagger$
              & 2021
              & 48.18 & 39.34 & \underline{62.15}
              & 54.44 & 42.60 & \textbf{75.40}
              & 46.96 & 43.28 & 51.34
              & 27.21 & 27.58 & 26.85
              & 32.07 & 20.39 & 75.14
              & 20.86 & 22.40 & 19.52
              & 13.29 & 8.84  & 26.82
              & 29.22 & 28.31 & 30.20 \\
            FR-UNet~\cite{FR-UNet} $\ddagger$
              & 2022
              & 42.01 & 32.53 & 59.28
              & 43.85 & 33.82 & 62.33
              & 50.08 & 40.11 & \underline{69.27}
              & 36.97 & 44.97 & 31.48
              & 17.51 & 9.86  & \underline{77.97}
              & 40.95 & 36.70 & 46.33
              & 26.91 & 30.03 & 24.38
              & 42.40 & 32.12 & 62.36 \\
            BCU-Net~\cite{BCU} $\ddagger$
              & 2023
              & 36.86 & 40.65 & 33.72
              & 39.94 & 52.35 & 32.29
              & 47.76 & 53.54 & 43.10
              & 36.36 & 53.39 & 27.57
              & 24.32 & 23.71 & 24.98
              & 32.58 & 42.12 & 26.57
              & 28.14 & 35.72 & 23.22
              & 42.37 & 35.55 & 52.42 \\
            \midrule
            SAM-Med2d~\cite{sam-med2d} ${\dagger}$
              & 2023
              & 49.29 & 46.67 & 52.23
              & 49.29 & 65.96 & 39.35
              & 52.91 & 61.97 & 46.16
              & \underline{44.02} & 60.03 & 34.75
              & 30.58 & \underline{45.91} & 22.93
              & 41.22 & \underline{59.29} & 31.59
              & 31.30 & 48.32 & 23.15
              & 45.02 & \underline{62.35} & 35.23 \\
            Rein~\cite{Regin} ${\dagger}$
              & 2024
              & 42.46 & 35.21 & 53.48
              & 52.07 & 54.36 & 49.97
              & 52.76 & 56.28 & 49.67
              & 40.07 & 33.14 & 50.67
              & 33.59 & 32.95 & 34.27
              & 47.21 & 42.26 & 53.47
              & 33.48 & 49.21 & 25.37
              & 41.95 & 32.41 & 59.46 \\
            HQ-SAM~\cite{SAMinHQ} ${\dagger}$
              & 2024
              & 49.01 & 43.74 & 55.73
              & 54.45 & 48.95 & 63.94
              & 51.99 & 42.72 & 66.39
              & 39.33 & 30.43 & 55.59
              & 34.91 & 23.88 & 64.91
              & 50.64 & 46.04 & 56.26
              & 42.20 & 37.20 & 48.76
              & 43.08 & 30.37 & 74.07 \\
            CWSAM~\cite{CWSAM} ${\dagger}$
              & 2025
              & \underline{50.10} & 44.50 & 56.80
              & \underline{55.60} & 49.80 & 65.10
              & \underline{53.10} & 43.60 & 67.20
              & 40.20 & 31.50 & \underline{56.80}
              & \underline{35.80} & 24.60 & 66.40
              & \underline{51.70} & 47.10 & \underline{57.40}
              & \underline{43.50} & 38.00 & 49.60
              & 44.00 & 31.20 & 75.10 \\
            KnowSAM~\cite{KnowSAM} ${\dagger}$
              & 2025
              & 47.50 & 42.00 & 54.00
              & 53.30 & 47.50 & 62.50
              & 50.20 & 41.80 & 65.50
              & 38.40 & 29.60 & 53.90
              & 34.00 & 22.90 & 63.80
              & 49.50 & 45.00 & 55.80
              & 41.00 & 36.50 & 46.50
              & 42.00 & 29.80 & 73.50 \\
            \midrule
            UCS (Ours)
              & ---
              & \textbf{65.59} & \textbf{64.63} & \textbf{66.59}
              & \textbf{69.31} & \textbf{73.78} & \underline{65.35}
              & \textbf{73.55} & \textbf{76.24} & \textbf{71.05}
              & \textbf{93.63} & \textbf{93.82} & \textbf{93.45}
              & \textbf{82.11} & \textbf{82.30} & \textbf{82.06}
              & \textbf{65.32} & \textbf{75.49} & \textbf{57.56}
              & \textbf{74.22} & \underline{69.47} & \textbf{79.69}
              & \textbf{85.46} & \textbf{93.11} & \textbf{78.97} \\
            \bottomrule
        \end{tabular}%
    }
\end{table*}

\begin{table*}[!t]
    \setlength{\tabcolsep}{3pt}
    \renewcommand{\arraystretch}{0.85}
    \centering
    \caption{Comparison results on the {\textbf{Public Test Dataset}}.}
    \label{tab: public data}
    \resizebox{\textwidth}{!}{%
    \begin{tabular}{l | l | ccc | ccc | ccc | ccc | ccc | ccc | ccc}
        \toprule
        \multirow{2}{*}{Method} & \multirow{2}{*}{Year}
            & \multicolumn{3}{c|}{OCTA500\_3M}
            & \multicolumn{3}{c|}{OCTA500\_6M}
            & \multicolumn{3}{c|}{CSD}
            & \multicolumn{3}{c|}{DCA1}
            & \multicolumn{3}{c|}{TubeTK}
            & \multicolumn{3}{c|}{STARE}
            & \multicolumn{3}{c}{FIVES} \\
        \cmidrule(lr){3-5} \cmidrule(lr){6-8} \cmidrule(lr){9-11} \cmidrule(lr){12-14} \cmidrule(lr){15-17} \cmidrule(lr){18-20} \cmidrule(lr){21-23}
        & & F1 $\uparrow$ & Pre $\uparrow$ & Rec $\uparrow$
          & F1 $\uparrow$ & Pre $\uparrow$ & Rec $\uparrow$
          & F1 $\uparrow$ & Pre $\uparrow$ & Rec $\uparrow$
          & F1 $\uparrow$ & Pre $\uparrow$ & Rec $\uparrow$
          & F1 $\uparrow$ & Pre $\uparrow$ & Rec $\uparrow$
          & F1 $\uparrow$ & Pre $\uparrow$ & Rec $\uparrow$
          & F1 $\uparrow$ & Pre $\uparrow$ & Rec $\uparrow$ \\
        \midrule
        UNet-vgg16~\cite{2015unet} $\ddagger$
            & 2015
            & 40.71 & 80.45 & 28.15
            & 53.23 & 79.05 & 41.67
            & 23.92 & 49.96 & 15.73
            & 47.12 & 72.05 & 35.44
            & 14.94 & 80.04 & 15.39
            & 47.45 & 87.70 & 32.23
            & 60.29 & 74.64 & 50.57 \\
        UNet-resnet101~\cite{2015unet} $\ddagger$
            & 2015
            & 30.26 & 56.35 & 30.12
            & 43.95 & 74.11 & 31.24
            & 18.79 & 40.61 & 12.23
            & 42.33 & 60.10 & 34.50
            & 3.95 & 46.20 & 2.90
            & 40.19 & 47.60 & 34.78
            & 22.41 & 49.33 & 14.50 \\
        UNet++~\cite{unet++} $\ddagger$
            & 2018
            & 75.57 & 71.15 & 78.54
            & 74.33 & 72.89 & 74.33
            & 30.71 & 45.64 & 22.27
            & 44.12 & 62.20 & 34.00
            & 35.44 & 60.26 & 29.34
            & 50.44 & 70.26 & 39.34
            & 58.13 & 80.65 & 45.89 \\    
        CSNet~\cite{csnet} $\ddagger$
            & 2019
            & 68.10 & 72.50 & 64.00
            & 64.00 & 78.00 & 55.00
            & 25.00 & 45.00 & 18.00
            & 55.05 & 60.30 & 50.20
            & 34.50 & 51.00 & 33.00
            & 45.50 & 61.00 & 38.00
            & 42.00 & 66.00 & 33.00 \\
        CS$^2$Net~\cite{cs2net} $\ddagger$
            & 2020
            & 69.60 & 73.00 & 65.50
            & 65.50 & 78.00 & 56.50
            & 24.50 & 44.50 & 16.50
            & 56.22 & 61.10 & 51.00
            & 34.00 & 50.50 & 32.50
            & 44.00 & 60.50 & 37.50
            & 39.50 & 63.50 & 30.50 \\
        CSSR~\cite{CSSR} $\ddagger$
            & 2021
            & 59.28 & 82.44 & 45.70
            & 70.92 & 63.53 & 80.05
            & 42.15 & 62.63 & 31.76
            & 62.45 & 70.10 & 55.90
            & 16.72 & 76.88 & \underline{74.40}
            & 11.72 & 66.88 & \underline{64.40}
            & 27.16 & 79.18 & 16.39 \\
        FR-UNet~\cite{FR-UNet} $\ddagger$
            & 2022
            & 76.23 & \textbf{82.78} & 68.70
            & 70.25 & \textbf{96.11} & 55.35
            & 35.67 & 45.11 & 29.50
            & 61.78 & 75.00 & 52.20
            & 16.53 & 55.90 & \textbf{76.70}
            & 11.53 & 66.70 & \textbf{66.70}
            & 59.71 & 67.29 & 55.63 \\
        BCU-Net~\cite{BCU} $\ddagger$
            & 2023
            & 69.44 & 70.08 & 68.81
            & 66.98 & 68.88 & 64.29
            & 22.11 & 54.16 & 12.83
            & 56.30 & 63.00 & 48.90
            & 41.89 & 70.79 & 33.91
            & 46.89 & 70.79 & 33.91
            & 34.16 & 73.70 & 19.99 \\
        \midrule
        SAM-Med2d~\cite{sam-med2d} ${\dagger}$
            & 2023
            & 72.38 & 72.04 & 72.73
            & 70.90 & 74.40 & 67.69
            & 41.10 & 49.85 & 34.69
            & 65.50 & 72.10 & 59.20
            & 49.99 & 73.67 & 43.14
            & 54.99 & 73.67 & 43.14
            & 60.24 & \underline{78.36} & 48.47 \\
        Rein~\cite{Regin} ${\dagger}$
            & 2024
            & 77.32 & 74.13 & 80.79
            & 77.83 & 75.20 & 80.64
            & 47.59 & 52.42 & \underline{43.58}
            & 71.80 & 77.00 & \textbf{67.50}
            & 31.30 & \underline{86.00} & 33.00
            & 26.30 & 76.00 & 23.00
            & 70.80 & 73.69 & 70.22 \\
        HQ-SAM~\cite{SAMinHQ} ${\dagger}$
            & 2024
            & 80.29 & 75.35 & 85.93
            & 78.68 & 76.74 & 80.73
            & 49.19 & 63.77 & 40.03
            & 73.20 & 82.00 & 65.00
            & 31.25 & \textbf{98.73} & 25.48
            & 26.25 & \textbf{88.73} & 15.48
            & 73.07 & 75.69 & 70.62 \\
        CWSAM~\cite{CWSAM} ${\dagger}$
            & 2025
            & \underline{81.20} & 76.20 & \textbf{86.80}
            & \underline{79.60} & 77.50 & \underline{81.50}
            & \underline{50.30} & \underline{64.80} & 41.20
            & \underline{74.10} & \underline{83.50} & \underline{66.50}
            & 47.88 & 82.42 & 40.63
            & 42.88 & 72.42 & 30.63
            & \underline{74.20} & 76.80 & \textbf{71.70} \\
        KnowSAM~\cite{KnowSAM} ${\dagger}$
            & 2025
            & 72.30 & 67.50 & 77.90
            & 70.70 & 69.00 & 72.90
            & 41.10 & 55.50 & 33.00
            & 64.40 & 72.20 & 57.70
            & \underline{52.20} & 68.80 & 48.60
            & \underline{57.20} & 68.80 & 48.60
            & 62.00 & 64.60 & 59.50 \\
        \midrule
        UCS (Ours)
            & ---
            & \textbf{84.45} & \underline{82.73} & \underline{86.23}
            & \textbf{82.48} & \underline{82.73} & \textbf{82.15}
            & \textbf{74.95} & \textbf{81.66} & \textbf{69.27}
            & \textbf{75.45} & \textbf{87.52} & 66.43
            & \textbf{57.59} & 62.73 & 62.15
            & \textbf{73.92} & \underline{88.61} & 63.43
            & \textbf{79.67} & \textbf{90.73} & \underline{71.02} \\
        \bottomrule
    \end{tabular}%
    }
\end{table*}

\subsubsection{Comparison on In-House Test Dataset}\label{open-seg}
The open-set segmentation performance was evaluated on the In-House Test Dataset, which includes a wide range of curvilinear structures from diverse real-world scenarios. 

Table~\ref{tab: our data} provides a comprehensive comparison of our proposed UCS method against several state-of-the-art segmentation models on the In-House Test Dataset, evaluating performance across eight diverse categories. Our UCS method demonstrates strong performance, achieving F1-score ranging from 65.32 (WIRE) to 93.63 (SCRATCH), significantly outperforming existing approaches. Across the categories listed from left to right in Table~\ref{tab: our data}, UCS outperforms the second-best comparison model by 16.30, 14.86, 20.64, 49.61, 47.20, 14.68, 32.02, and 39.44 on the F1-score, respectively. For example, in the CRACK category, UCS achieves an F1-score of 69.31, surpassing HQ-SAM's F1-score of 54.45 by a notable margin of 14.86 points. These results underscore UCS's exceptional generalization capability, establishing it as a new benchmark for curvilinear structure segmentation tasks.

\subsubsection{Comparison on Public Test Dataset}\label{general-seg}
To verify the consistency of performance gains, we conducted experiments on the Public Test Dataset.

Table \ref{tab: public data} compares UCS with other segmentation models on the Public Test dataset. UCS attains the highest F1-score on all seven public benchmarks (OCTA500\_3M, OCTA500\_6M, CSD, DCA1, TubeTK, STARE, and FIVES). Although UCS does not always achieve the top precision or recall individually (e.g., FR‑UNet reports higher precision on OCTA500\_3M and OCTA500\_6M), this is a consequence of different models operating at different points on the precision–recall spectrum. Some methods over-segment and therefore obtain high recall but lower precision, while others produce conservative, highly precise masks that miss thin or fragmented structures and thus have lower recall. The F1-score, as the harmonic mean of precision and recall, captures the balance between these behaviors. UCS is deliberately biased toward recovering fine, continuous curvilinear structures—accepting a modest rise in false positives to secure higher recall and better structural coherence—so its precision may not always be maximal, but its overall F1 is consistently superior. The particularly large margin on CSD (~24.7 F1 points over the next best) further underscores UCS's effectiveness on crack-like patterns and its strong cross-domain generalisation.

Fig.~\ref{fig: visualization compare} presents a visual comparison of segmentation results across a range of datasets. In the FIVES dataset, our method clearly outperforms others, exhibiting precise delineation of vascular structures with minimal red (false negatives) and green (false positives) pixels. In the LEAF dataset, our approach segments the delicate vein patterns, while other methods produce incomplete or fragmented results, primarily showing red pixels indicating significant false negatives. In the SOIL dataset, other methods exhibit significant noise and incomplete segmentation. Our method provides a complete and clean segmentation. Overall, the proposed UCS consistently outperforms competing methods by delivering higher true positive rates, lower false positives, and superior structural coherence across all examined domains.

\setlength{\tabcolsep}{0.5pt}
\begin{figure*}[t]
\centering
\begin{tabularx}{\textwidth}{
    >{\raggedleft\arraybackslash}m{0.3cm} 
    *{9}{>{\centering\arraybackslash}m{0.106\textwidth}}
}
\rotatebox[origin=c]{90}{\raisebox{2.0ex}{\parbox[l][0.3cm]{1.0cm}{\centering \textbf{LEAF$^{\ast}$}}}} &  

\includegraphics[width=\linewidth, height=5cm, keepaspectratio]{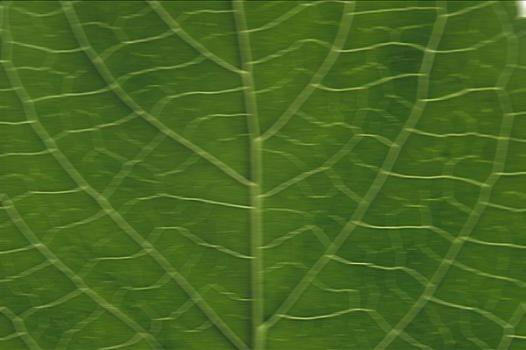} &
\includegraphics[width=\linewidth, height=5cm, keepaspectratio]{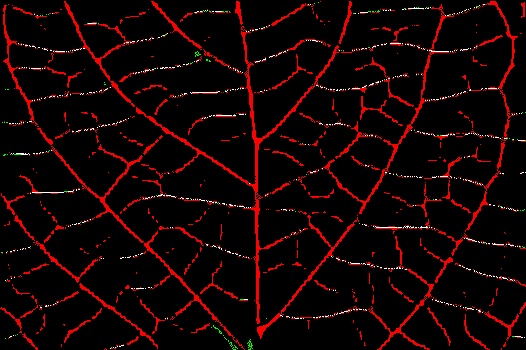} & 
\includegraphics[width=\linewidth, height=5cm, keepaspectratio]{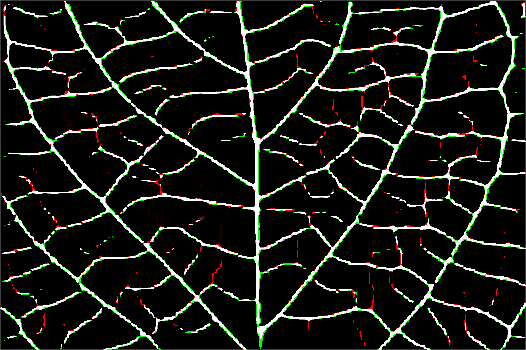} & 

\includegraphics[width=\linewidth, height=5cm, keepaspectratio]{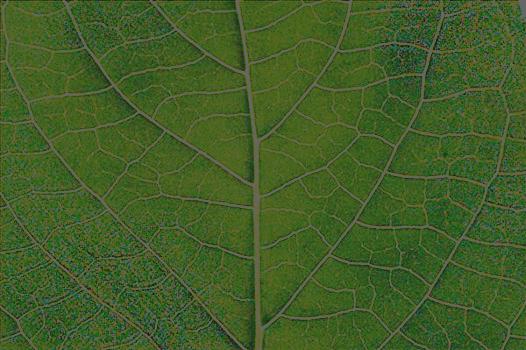} &
\includegraphics[width=\linewidth, height=5cm, keepaspectratio]{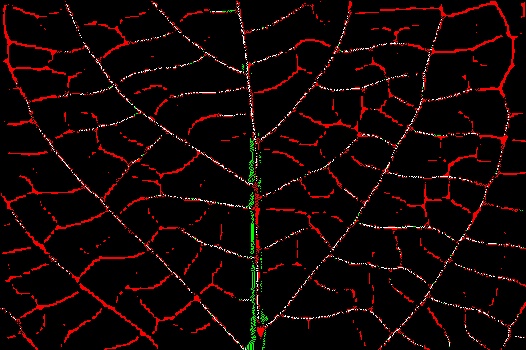} & 
\includegraphics[width=\linewidth, height=5cm, keepaspectratio]{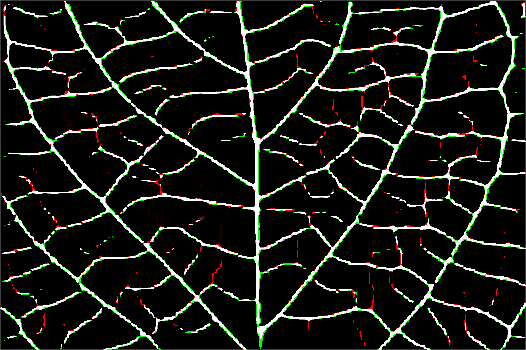} & 

\includegraphics[width=\linewidth, height=5cm, keepaspectratio]{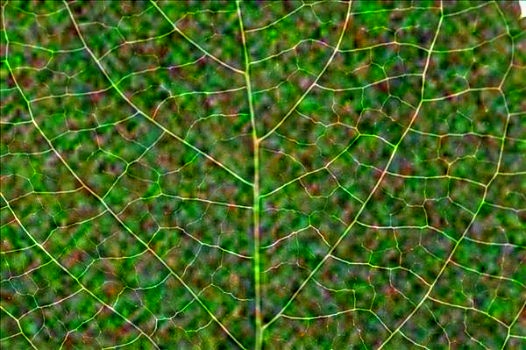} &
\includegraphics[width=\linewidth, height=5cm, keepaspectratio]{noise_show/frunet/danet/leaf_34.jpg} &
\includegraphics[width=\linewidth, height=5cm, keepaspectratio]{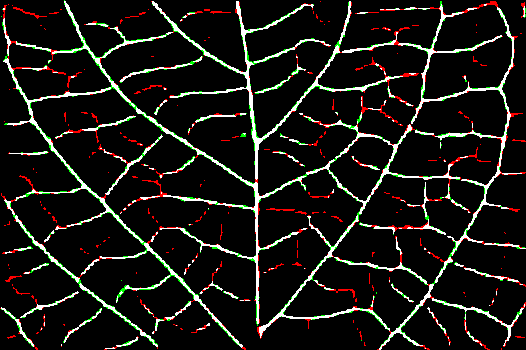} \\[-2.3pt]

\rotatebox[origin=c]{90}{\raisebox{-1.0ex}{\parbox[l][0.3cm]{1.0cm}{\centering \textbf{\small OCTA\_3M}}}} &

\includegraphics[width=\linewidth, height=4cm, keepaspectratio]{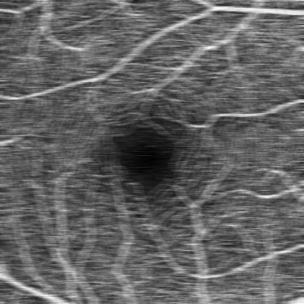} &
\includegraphics[width=\linewidth, height=4cm, keepaspectratio]{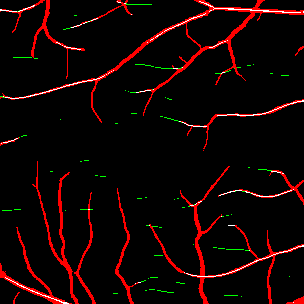} & 
\includegraphics[width=\linewidth, height=3cm, keepaspectratio]{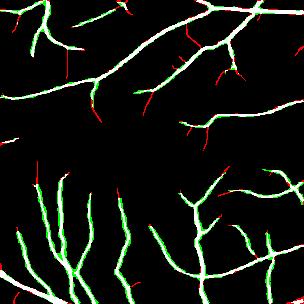} & 

\includegraphics[width=\linewidth, height=4cm, keepaspectratio]{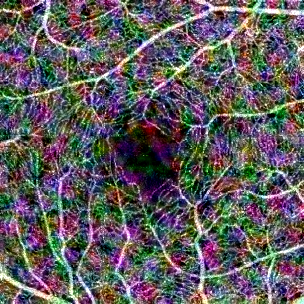} &
\includegraphics[width=\linewidth, height=4cm, keepaspectratio]{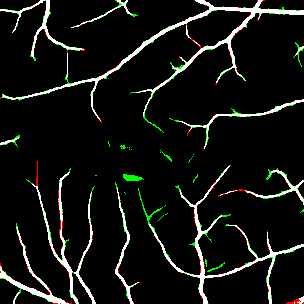} &
\includegraphics[width=\linewidth, height=4cm, keepaspectratio]{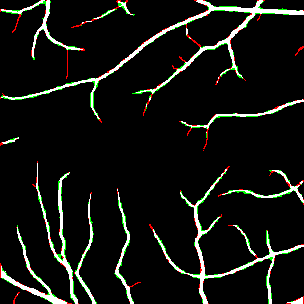} &

\includegraphics[width=\linewidth, height=4cm, keepaspectratio]{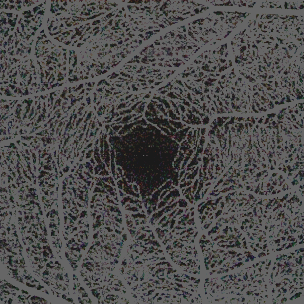} & 
\includegraphics[width=\linewidth, height=4cm, keepaspectratio]{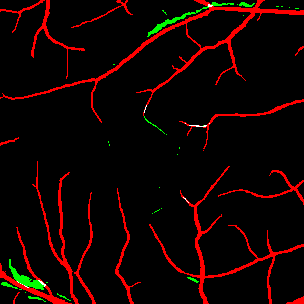} & 
\includegraphics[width=\linewidth, height=4cm, keepaspectratio]{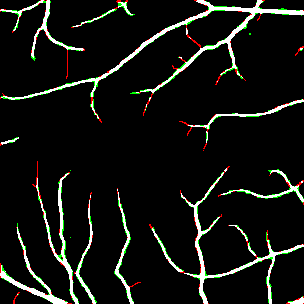} \\[-8pt]
\multicolumn{1}{c}{} &
\multirow{2}{*}{\textbf{\small Motion}} & \multirow{2}{*}{\textbf{\small FR-UNet~\cite{FR-UNet}}} & \multirow{2}{*}{\textbf{\small Ours}} & \multirow{2}{*}{\textbf{\small sRNS~\cite{sRGB}}} & \multirow{2}{*}{\textbf{\small FR-UNet~\cite{FR-UNet}}} & \multirow{2}{*}{\textbf{\small Ours}} & \multirow{2}{*}{\textbf{\small DANET~\cite{DANET}}} & \multirow{2}{*}{\textbf{\small FR-UNet~\cite{FR-UNet}}} & \multirow{2}{*}{\textbf{\small Ours}}\\[5pt]
\end{tabularx}
\caption{Visualization of image degradation and segmentation results post-degradation. Columns 1, 4, and 7 show the results of the degradation methods indicated by the text at the bottom, with the segmentation results of FR-UNet and UCS (Ours) displayed to their right.}
\label{fig: noisecompare}
\end{figure*}

\subsection{Robustness Study}\label{robust}
We evaluated UCS's robustness by testing its segmentation performance under simple and realistic image degradations.

\subsubsection{Simple Degradations} 
Gaussian noise is random fluctuations following a normal distribution and is added to the input images with a zero mean and varying standard deviations ($\sigma$ =10, 15). The noisy image is defined as:
\begin{equation}
I_{y} = I + N(0,\sigma ^{2} )
\end{equation}

Motion Blur refers to image degradation caused by relative motion between the camera and the scene during exposure, typically simulated by convolving the image with a motion blur kernel. The kernel is defined as:
\begin{equation}
K(L, \phi) = 
\begin{cases} 
\frac{1}{L} & \text{if } \sqrt{x^2 + y^2} \leq \frac{L}{2} \text{ and } \frac{y}{x} = \tan(\phi), \\
0 & \text{otherwise}.
\end{cases}
\end{equation} 
where $x,y$ are the coordinates in the kernel, $L$ denotes the kernel dimensions, which are set to 7, 11, and 15 in this experiment, \(\phi \in [0, \pi)\) represents the motion direction, and $\phi$ is randomly assigned.

\subsubsection{Realistic Degradations}
We employed three advanced degradation models: sRGB-Real-Noise-Synthesizing (sRNS)~\cite{sRGB}, DASR~\cite{DASR}, and DANET~\cite{DANET}. sRNS improves the realism of degradation by modeling spatial correlations and signal dependencies through three specialized networks. DASR predicts degradation parameters such as Gaussian blur, downsampling, Poisson noise, and JPEG compression through a lightweight regression network, which dynamically simulates realistic degradation effects, including blurring and noise artifacts. DANET generates realistic degradations by learning the joint distribution of clean and degraded images via adversarial training, effectively capturing both spatial correlations and intensity variations.

\begin{table}[t]
\centering
\caption{Performance under various degeneration on the Public Test Dataset (Public) and the In-House Test Dataset (In-House).}
\label{tab: robustness_results}
    \begin{tabularx}{0.95\columnwidth}{l|*{3}{>{\centering\arraybackslash}X}|*{3}{>{\centering\arraybackslash}X}}
        \toprule
        \multirow{2}{*}{\centering Degradation} & \multicolumn{3}{c|}{Public} & \multicolumn{3}{c}{In-House}\\
        \cmidrule(lr){2-4} \cmidrule(lr){5-7} 
        & F1 $\uparrow$ & Pre $\uparrow$ & Rec $\uparrow$ & F1 $\uparrow$ & Pre $\uparrow$ & Rec $\uparrow$\\ \midrule
        \multicolumn{1}{c|}{-} & 77.74 & 84.51 & 71.27 & 74.69 & 78.61 & 71.16 \\
        \midrule
        Gaussian($\sigma$=10) & 76.55 & 83.19 & 70.91 & 74.05 & 76.72 & 71.56 \\
        Gaussian($\sigma$=15)& 75.30 & 81.52 & 69.97 & 73.37 & 76.19 & 70.75 \\ 
        Motion(L=7)& 74.28 & 78.61 & 70.40 & 73.13 & 72.49 & 73.80 \\
        Motion(L=11)& 73.49 & 76.70 & 70.53 & 71.98 & 70.49 & 73.54 \\
        Motion(L=15)& 70.37 & 75.04 & 66.25 & 70.29 & 68.20 & 72.52 \\
        \midrule
        sRNS~\cite{sRGB}& 67.75 & 84.26 & 56.59 & 70.09 & 70.02 & 70.16 \\
        DASR~\cite{DASR}& 75.15 & 80.49 & 70.47 & 70.68 & 68.50 & 73.02 \\
        DANET~\cite{DANET}& 72.68 & 80.80 & 66.04 & 70.74 & 70.92 & 70.57 \\ 
        \bottomrule
    \end{tabularx}
\end{table}

As shown in Table~\ref{tab: robustness_results}, UCS consistently achieves high segmentation accuracy across various degradation methods and levels. Gaussian noise with $\sigma$=10 and $\sigma$=15 leads to a slight decline in the F1-score, dropping from 77.74 to 76.55 and 75.30 on the Public test dataset, and from 74.69 to 74.05 and 73.37 on the In-House test dataset. Motion blur with increasing lengths (L=7, L=11, L=15) further impacts performance, with the F1-score on the Public test dataset reducing to 74.28, 73.49, and 70.37, respectively, and on the In-House test dataset declining to 73.13, 71.98, and 70.29. Among deep learning-based degradation models, sRNS causes the most significant F1-score drop, with scores decreasing to 67.75 and 70.09 on both two major datasets. DASR and DANET lead to moderate performance drop, with F1-score of 75.15 and 72.68 on the Public test dataset, and 70.76 and 70.74 on the In-House test dataset, respectively.

Fig.~\ref{fig: noisecompare} provides a qualitative comparison of segmentation results under various image degradations. We observe that UCS consistently outperforms FR-UNet in handling degraded images. For example, under Motion blur, FR-UNet fails to capture the leaf veins, resulting in fragmented segmentation with many red pixels (false negatives), whereas UCS accurately segments the veins. Similarly, with sRNS noise, FR-UNet exhibits numerous green pixels (false positives), while UCS maintains accurate segmentation. Under DANet degradation, FR-UNet produces significant red and green errors, struggling to preserve details, while UCS delivers sharper segmentation. These results demonstrate the superior robustness of UCS against diverse image degradations.

\begin{table}[t]
    \centering 
    \caption{Ablation of sa, pg, hfc, and gfc modules.}
    \label{tab: ablation}
    \begin{tabularx}{0.95\columnwidth}{l|c@{\hspace{0.15cm}}c@{\hspace{0.15cm}}c@{\hspace{0.15cm}}c|*{3}{>{\centering\arraybackslash}X}|*{3}{>{\centering\arraybackslash}X}}
        \toprule
        \multirow{2}{*}{\centering Model } & \multicolumn{4}{c|}{Module} & \multicolumn{3}{c|}{Public} & \multicolumn{3}{c}{In-House}\\
    \cmidrule(lr){2-5} \cmidrule(lr){6-8} \cmidrule(lr){9-11}
        & SA & PG & HFC & GFC & F1 $\uparrow$ & Pre $\uparrow$ & Rec $\uparrow$ & F1 $\uparrow$ & Pre $\uparrow$ & Rec $\uparrow$ \\ \midrule
        & - &  - &  - &  - & 33.07 & 39.57 & 28.41 & 29.95 & 31.87 & 28.25 \\
        & \checkmark &  - &  - &  - & 45.45 & 65.93 & 36.80 & 40.92 & 57.87 & 31.65 \\
        Base & \checkmark  & \checkmark & -  & -  & 61.57 & 82.53 & 49.10 & 59.54 & 72.99 & 50.28 \\ 
        & \checkmark & \checkmark & \checkmark & -  & 67.78 & 81.19 & 58.18 & 66.35 & 76.00 & 58.87\\
        & \checkmark & \checkmark &  \checkmark & \checkmark & 77.74 & 84.51 & 71.27 & 74.69 & 78.61 & 71.16 \\
        \bottomrule
    \end{tabularx}
\end{table}

\begin{figure}[t]
    \centering
    \includegraphics[width=0.9\linewidth]{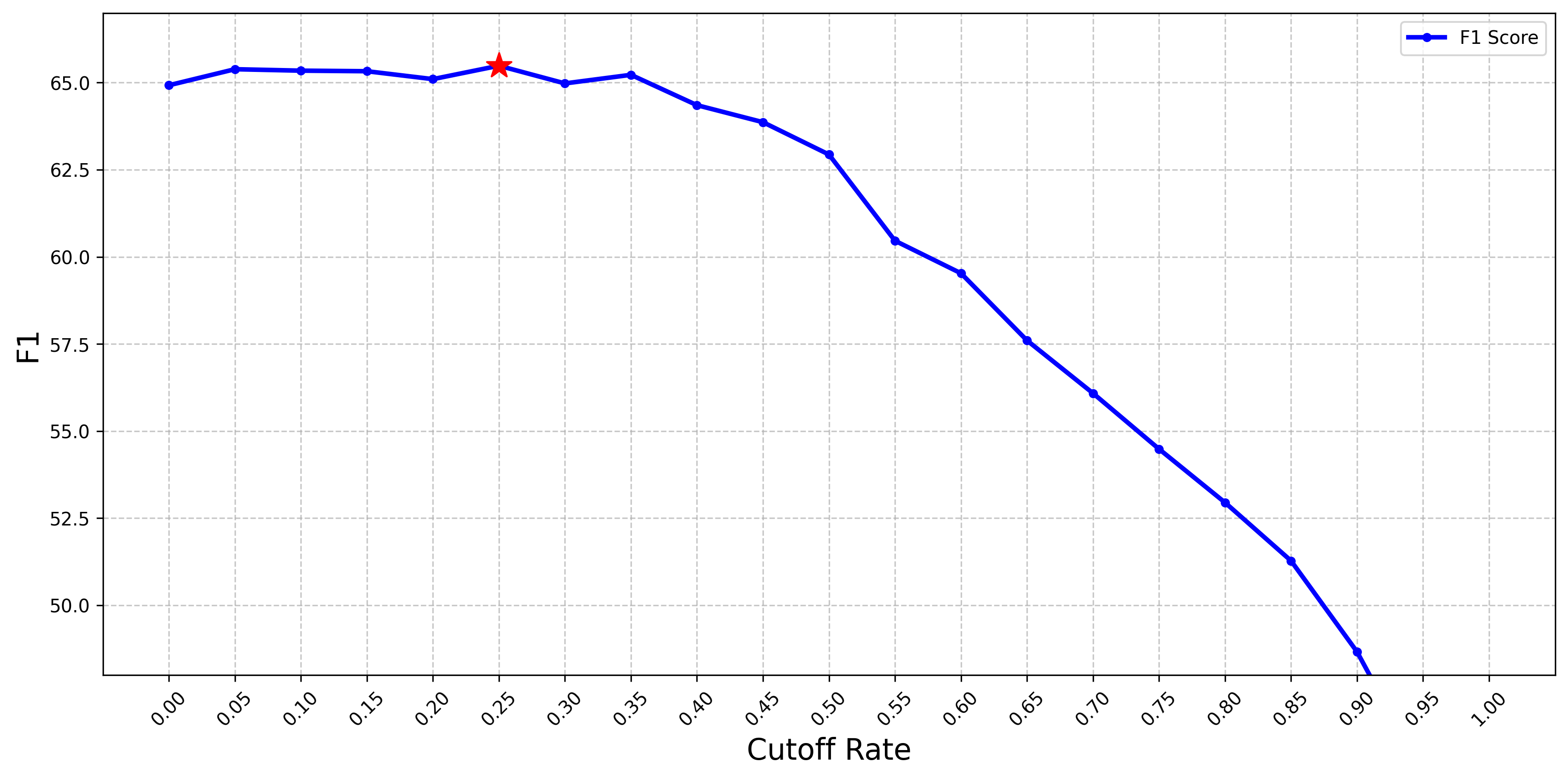}
    \caption{The impact of different cutoff rates on the F1 on WIRE datasets. Performance peaks at a cutoff rate of 0.25 and subsequently declines.}
    \label{fig:cutoff_rate}
\end{figure}

\begin{table}[t]
\centering
\caption{Performance comparison of different adapter sparse strategies in the encoder blocks, where - denotes the original SAM model.}
\label{tab:adapter_comp}
\begin{tabularx}{0.95\columnwidth}{>{\centering\arraybackslash}p{1.85cm}|*{3}{>{\centering\arraybackslash}X}|*{3}{>{\centering\arraybackslash}X}}
    \toprule
    \multirow{2}{*}{Sparse Adapter} & \multicolumn{3}{c|}{Public} & \multicolumn{3}{c}{In-House}\\
    \cmidrule(lr){2-4}\cmidrule(lr){5-7}
      Number  & F1 $\uparrow$ & Pre $\uparrow$ & Rec $\uparrow$ & F1 $\uparrow$ & Pre $\uparrow$ & Rec $\uparrow$ \\ \midrule
    - & 33.07 & 39.57 & 28.41 & 29.95 & 31.87 & 28.25 \\
    24 & 33.74 & 42.50 & 27.98 & 27.56 & 28.13 & 27.02 \\
    12 & 34.75 & 45.03 & 28.29 & 29.66 & 31.26 & 28.22 \\
    8 & 34.28 & 48.98 & 26.34 & 35.86 & 48.67 & 28.39 \\
    6 & 36.72 & 50.18 & 28.95 & 38.20 & 53.73 & 29.63 \\
    4 & \textbf{45.45} & \textbf{65.93} & \textbf{36.80} & \textbf{40.92} & \textbf{57.87} & \textbf{31.65} \\
    3 & 30.79 & 48.69 & 22.51 & 26.33 & 36.81 & 20.49 \\
    \bottomrule
\end{tabularx}
\end{table}

\subsection{Ablation Study}
\subsubsection{The Impact of Different Components}
Table~\ref{tab: ablation} presents a comprehensive ablation study, assessing the contributions of each module within the UCS framework. The analysis reveals a progressive performance enhancement with the sequential integration of each component. Specifically, the SA module demonstrably improves F1-score by 12.36 and 10.97 on the public and in-house datasets, respectively. Subsequent inclusion of the PG module further augments performance, yielding precision gains of 16.60 and 15.12, alongside recall improvements of 12.30 and 18.63. The HFC module leads to a consistent increase in recall, with gains of 9.08 and 8.59 observed across both datasets. Finally, the GFC module exhibits a pronounced impact on recall, achieving increments of 13.09 and 12.19, while exhibiting modest precision gains of 3.32 and 2.61. In summation, these results illustrate the incremental and complementary contributions of each UCS module.

\subsubsection{The Impact of Cutoff Rate}
To determine the optimal cutoff rate for the Prompt Generation (PG) module, we conducted an ablation study using a grid search approach. As illustrated in Fig.~\ref{fig:cutoff_rate}, the F1 score peaks when the cutoff rate is set to 0.25. As the rate increases beyond this point, performance begins to decline. This trend suggests that a lower cutoff rate fails to adequately filter out low-frequency noise, whereas a higher rate may inadvertently remove essential high-frequency structural details along with the noise. Therefore, a cutoff rate of 0.25 strikes the optimal balance, maximizing the extraction of relevant curvilinear features while minimizing interference from irrelevant information.

\begin{figure}[t]
    \centering
    \includegraphics[width=0.999\linewidth]{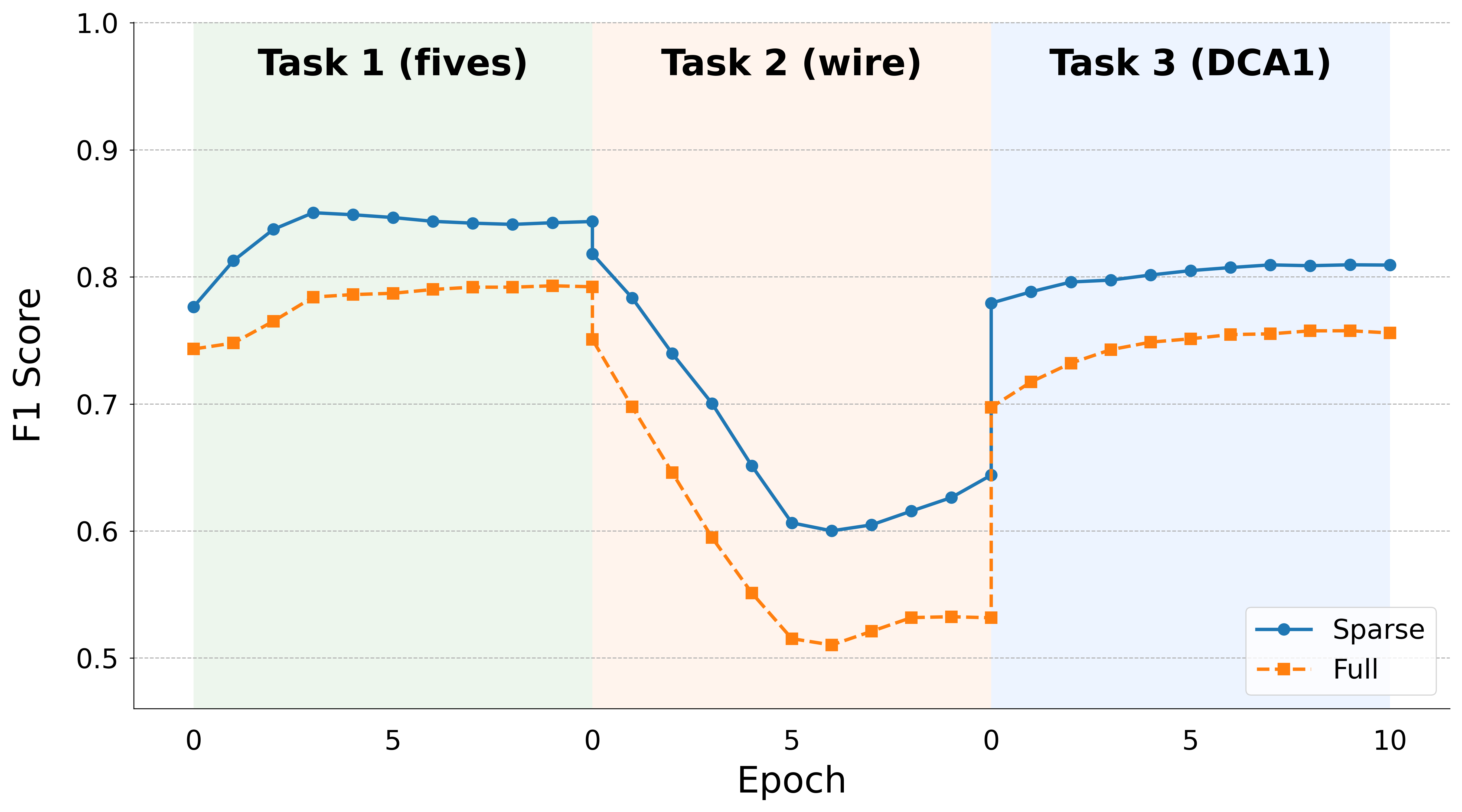}
    \caption{The F1 of the sparse adapter versus adapters inserted in all encoder blocks across three-stage training epochs.}
    \label{sparse}
\end{figure}
\subsubsection{Impact of Sparse Adapter Strategy}
Table~\ref{tab:adapter_comp} shows that the 4-adapter sparse configuration yields the best trade-off between performance and parameter efficiency: it improves F1 by 12.38 points on the public dataset and by 10.97 points on the in-house dataset versus the original SAM baseline. Overly dense adaptation (24 adapters) degrades performance, while extreme sparsity (3 adapters) also harms results, indicating a sweet spot for the number of adapters.

To verify the effectiveness of the Sparse Adapter strategy in mitigating catastrophic forgetting during domain shifts, we conducted a sequential training experiment. The model was trained first on FIVES (Task 1), then fine-tuned on Wire (Task 2), and finally on DCA1 (Task 3). As illustrated in Fig.~\ref{sparse}, both strategies exhibit similar evolutionary trends across the training epochs. However, the Sparse Adapter consistently outperforms the full insertion strategy, maintaining higher F1 scores throughout all stages. Specifically, when shifting to the second task, although both curves show a decline due to domain differences, the Sparse Adapter exhibits a much milder drop and superior stability. This demonstrates that by confining parameter updates to a smaller subspace, the sparse strategy reduces the risk of overfitting and retains better generalization capabilities compared to inserting adapters in every block.

\begin{figure}[t]
 \centering
 \subfloat[]{\includegraphics[width=0.49\columnwidth]{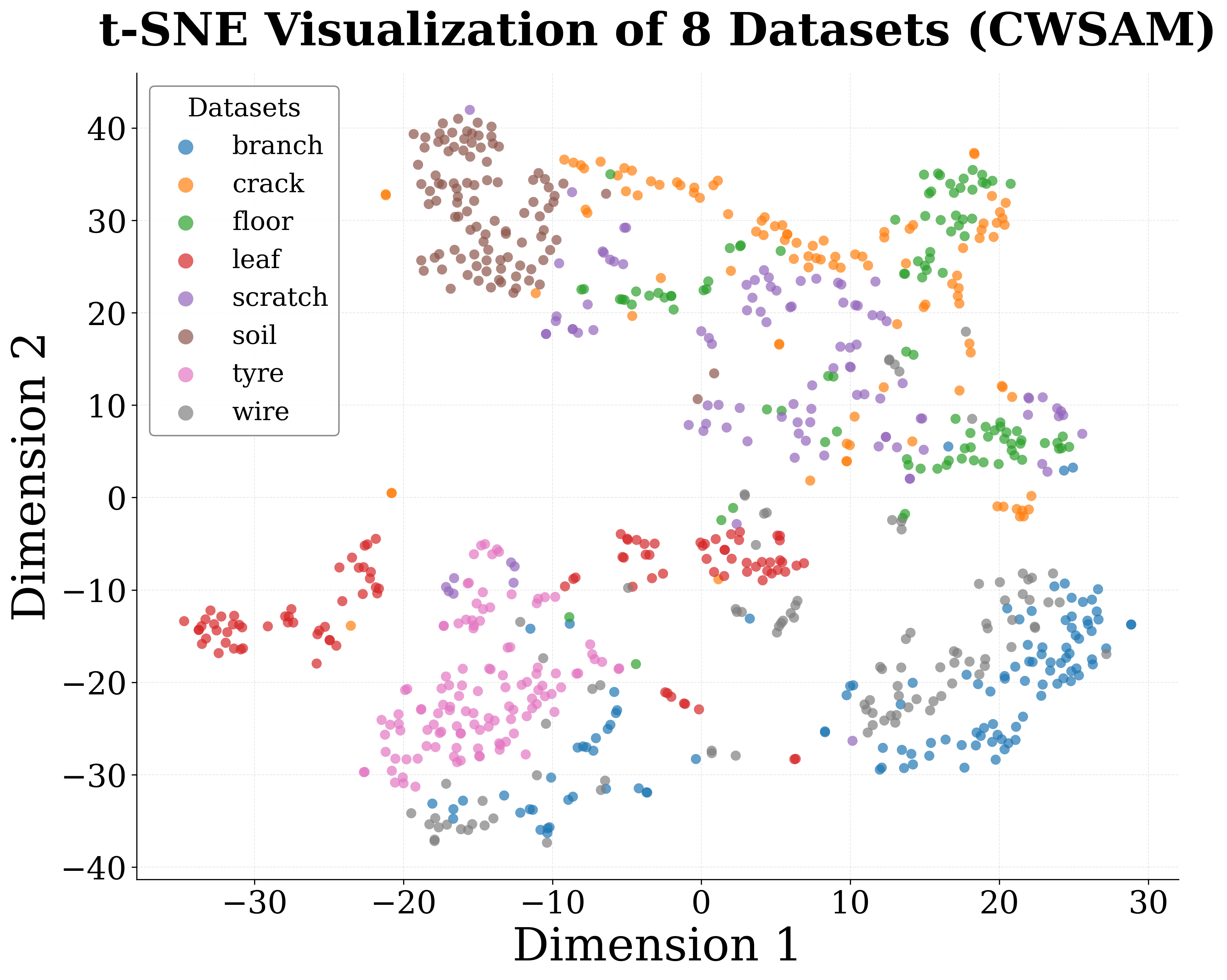}}
 \hfill
 \subfloat[]{\includegraphics[width=0.49\columnwidth]{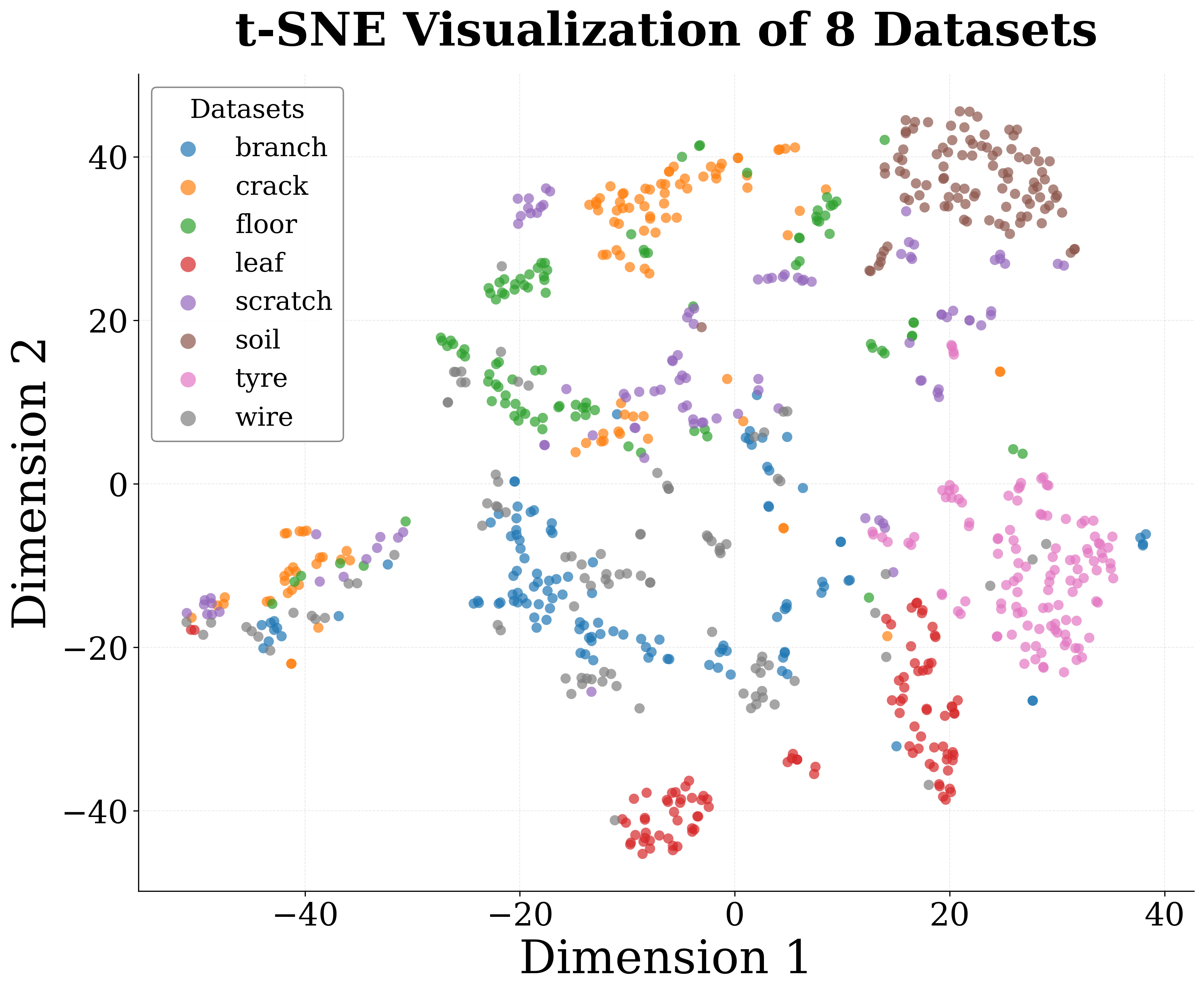}}
 \caption{Visualization of t-SNE embeddings for (a) CWSAM and (b) UCS on the eight In-House Test Dataset. Different colors represent different curvilinear structure categories.}
 \label{t-sne}
\end{figure}

\subsubsection{Visualization of t-SNE}
Figure~\ref{t-sne} illustrates the t-SNE embeddings derived from the In-House Test Dataset, comparing CWSAM (a) with UCS (b). A visual comparison reveals that UCS generates significantly more compact and distinctly separated clusters across different domains than CWSAM. For instance, in the CWSAM visualization (a), the green ('floor') and orange ('crack') clusters exhibit notable overlap, indicating poor discrimination between these categories. In stark contrast, the UCS visualization (b) demonstrates a clear separation of these same green and orange clusters, with each forming a more cohesive group. Furthermore, other clusters, such as red ('leaf') and pink ('scratch'), also show improved compactness and reduced inter-mixing in the UCS plot. This enhanced separation and compactness signify an improved inter-domain discrimination and more consistent semantic representations for various curvilinear structure categories with UCS, which directly correlates with its superior cross-domain segmentation performance.

\begin{table}[t]
  \centering
  \caption{Trainable parameter fraction, F1 and efficiency comparison of adaptation methods on WIRE dataset.}
  \label{trainable}
  \footnotesize
  \resizebox{\columnwidth}{!}{%
  \begin{tabular}{l c c c c c}
    \toprule
    Method & Trainable Params(\%)$\downarrow$ & F1$\uparrow$ & Params(M)$\downarrow$ & FLOPs(GB)$\downarrow$ & Runtime(ms)$\downarrow$ \\
    \midrule
    SAM-Med2d  & 68.0          & 41.22          & 950.14 & 1690.34 & 538.32 \\
    HQ-SAM     & \textbf{1.30} & 50.64          &\textbf{312.34} & 1493.84 & 268.46 \\
    CWSAM      & 12.1          & 51.70          & 315.56 & \textbf{1468.74} & 350.51 \\
    UCS (Ours) & 2.65          & \textbf{65.32} & 312.61 & 1497.88 & \textbf{239.56} \\
    \bottomrule
  \end{tabular}%
  }
\end{table}

\subsubsection{Number of Trainable Parameters}
The sparse-adapter design updates only 2.65\% of the model parameters by inserting lightweight adapters into four selected encoder blocks while keeping the rest of the pretrained encoder frozen. Paired with the Prompt Generation, Hierarchical Feature Compression and Guidance Feature Compression modules, this parameter-efficient adaptation preserves pretrained generalization and achieves the best empirical segmentation performance (e.g., F1 = 65.32) on our benchmarks. As shown in Table~\ref{trainable}, the proposed approach attains the highest F1 while maintaining a very low fraction of trainable parameters, demonstrating superior parameter efficiency without sacrificing accuracy.

\section{Conclusion}
In this paper, we introduce UCS, a universal segmentation model designed to segment arbitrary curvilinear structures across diverse domains, effectively bridging the domain gap. UCS features a novel architecture that integrates initial prompt information into each encoder block and fuses multi-level contextual features within the mask decoder, achieving a balance between global semantic understanding and precise local delineation. Comprehensive evaluations on both public and in-house test datasets demonstrate that UCS outperforms state-of-the-art CNN-based and Transformer-based models in segmentation accuracy, while also exhibiting robust performance under various image degradation conditions. By focusing on cross-domain segmentation, UCS transcends specific downstream tasks, aligning with our goal of universal applicability. In the future, we will focus on further reducing training parameters to achieve efficient model deployment and lower computational costs and explore fusing multi-modal data to improve segmentation accuracy in complex scenarios.


\bibliographystyle{IEEEtran}
\bibliography{myrefs}
\end{document}